\documentclass[twoside,11pt]{article}

\makeatletter
\let\Ginclude@graphics\@org@Ginclude@graphics
\makeatother

\usepackage[preprint]{jmlr2e}

\usepackage{microtype}
\usepackage{graphicx}
\usepackage{subfigure}
\usepackage{booktabs} 

\usepackage{epsfig,epsf,fancybox}
\usepackage{amsmath}
\usepackage{mathrsfs}
\usepackage{amssymb}
\usepackage{amsfonts}
\usepackage{graphicx}
\usepackage{color}
\usepackage{stmaryrd}
\usepackage{multirow}
\usepackage{booktabs}
\usepackage{float}
\usepackage{bbm}

\usepackage{algorithm}
\usepackage{algorithmic}

\def\proof{\par\noindent{\em Proof.}}

\usepackage{mathtools}

\usepackage[normalem]{ulem} 

\newcommand{\bm}[1]{\boldsymbol{#1}}

\newcommand{\R}{\mathbb{R}}

\def\bfe{{\mathbf e}}

\def\bfv{{\mathbf v}}
\def\bfw{{\mathbf w}}
\def\bfx{{\mathbf x}}
\def\bfy{{\mathbf y}}
\def\bfz{{\mathbf z}}

\def\bfR{{\mathbf R}}

\def\mcM{{M}}

\newcommand{\mbR}{\mathbb{R}}
\newcommand{\mcN}{\mathcal{N}}
\newcommand{\mbP}{\mathbb{P}}

\newcommand{\x}{\mathbf{x}}
\newcommand{\y}{\mathbf{y}}
\newcommand{\z}{\mathbf{z}}

\newcommand{\e}{\mathbf{e}}

\newcommand{\bc}{\begin{center}}
	\newcommand{\ec}{\end{center}}

\newcommand{\bdm}{\begin{displaymath}}
	\newcommand{\edm}{\end{displaymath}}

\newcommand{\beq}{\begin{equation}}
	\newcommand{\eeq}{\end{equation}}

	\newcommand{\efl}{\end{flushleft}}

\newcommand{\bt}{\begin{tabbing}}
	\newcommand{\et}{\end{tabbing}}

\newcommand{\beqn}{\begin{eqnarray}}
	\newcommand{\eeqn}{\end{eqnarray}}

\newcommand{\beqs}{\begin{align*}} 
	\newcommand{\eeqs}{\end{align*}}  

\begin{document}
	
	\title{Outlier Detection Using Generative Models with Theoretical Performance Guarantees}
	
	\author{Jirong Yi*, Jingchao Gao*, Tianming Wang, Xiaodong Wu, Weiyu Xu
\thanks{Jirong Yi, Jingchao Gao, Tianming Wang, Xiaodong Wu and Weiyu Xu are with the University of Iowa.}
\thanks{* contribute equally.}}
	
\editor{}
	
	\maketitle
\begin{abstract}
This paper considers the problem of recovering signals modeled by generative models from linear measurements contaminated with sparse outliers. We propose an outlier detection approach for reconstructing the ground-truth signals modeled by generative models under sparse outliers. We establish theoretical recovery guarantees for reconstruction of signals using generative models in the presence of outliers, giving lower bounds on the number of correctable outliers. Our results are applicable to both linear generator neural networks and the nonlinear generator neural networks with an arbitrary number of layers. We propose an iterative alternating direction method of multipliers (ADMM) algorithm for solving the outlier detection problem via $\ell_1$ norm minimization, and a gradient descent algorithm for solving the outlier detection problem via squared $\ell_1$ norm minimization. We conduct extensive experiments using variational auto-encoder and deep convolutional generative adversarial networks, and the experimental results show that the signals can be successfully reconstructed under outliers using our approach. Our approach outperforms the traditional Lasso and $\ell_2$ minimization approach.

\end{abstract}


\section{Introduction}
Recovering signals from its compressed measurements, also called compressed sensing (CS), has found applications in many fields, such as image processing, matrix completion and astronomy \cite{CandesRombergTao:2004,Donoho:2006:1,Candes:2009,Lustig:2007,Jaspan:2015,cai_robust_2016,bobin_compressed_2008,xu_separation-free_2018}. In some {{applications}}, faulty sensors and malfunctioning measuring system can give us measurements contaminated with outliers or gross errors \cite{mitra_analysis_2013,carrillo_robust_2010,studer_recovery_2012,xu_sparse_2013}, {{leading to}} the problem of recovering the true signal and detecting the {{outliers}} from the mixture of them. Specifically, suppose the true signal is $\bfx\in\mathbb{R}^n$, and the measurement matrix is $\mcM\in\mathbb{R}^{m\times n}$, then the measurement will be
  $\y = {\mcM}\x + \e + \bm{\eta}$, 
  where $\e\in\mathbb{R}^{m}$ is an $l$-sparse outlier {{vector}} due to sensor failure or system malfunctioning, and $l$ is usually much {{smaller}} than ${m}$, and the $\bm{\eta}\in\mbR^m$ is a noise vector. Our goal is {{to recover}} the true signal $\x$ and detect $\e$ from the measurement $\y$, and we call it the outlier detection problem. {{The outlier detection problem has attracted}} the interests from different fields, such as system identification, channel coding and image and video processing \cite{Bai:2017,xu_sparse_2013,barner_nonlinear_2003,CandesTao:2005,wright_dense_2010}.
 
There exists a large volume of works on the outlier detection problem, ranging from the practical algorithms for recovering the true signal and detecting the outliers \cite{CandesTao:2005,wright_dense_2010,xu_sparse_2013,wan_robust_2017,popilka_signal_2007}, to theoretical recovery guarantees \cite{CandesTao:2005,wright_dense_2010,xu_sparse_2013,studer_recovery_2012,popilka_signal_2007,mitra_analysis_2013,carrillo_robust_2010,candes_highly_2008}.
  {{
  In \cite{CandesTao:2005}, the channel decoding problem is considered and {{the authors proposed $\ell_{1}$ minimization for {{outlier}} detection under linear measurements. By assuming {{$m>n$}}, {{Candes and Tao}} found an annihilator to transform the error correction problem into a basis pursuit problem, and established the {{theoretical}} recovery guarantees. Later in \cite{wright_dense_2010}, Wright and Ma considered the same problem in the case where $m< n$. By assuming the true signal $\x$ is sparse, they reformulated the problem as an $\ell_1/\ell_1$ minimization. Based on a set of assumptions on the measuring matrix $\mcM$ and the sparse signals $\x$ and $\e$, they also established {{theoretical}} recovery guarantees under sparse outliers. In \cite{xu_sparse_2013}, the authors studied the sparse error correction problem with nonlinear measurements, {{and proposed {{an}} iterative $\ell_1$ minimization to detect outliers}}. By using a local linearization technique and an iterative $\ell_1$ minimization algorithm, they showed that with high probability, the true signal could be  recovered to high precision, and the iterative $\ell_1$ minimization algorithm would converge to the true signal. {{Other works on outlier detection include but are not limited to \cite{candes_decoding_2005,dwork_price_2007,popilka_signal_2007,candes_highly_2008,carrillo_robust_2010,wright_dense_2010,studer_recovery_2012,xu_sparse_2013,foygel_corrupted_2014,bhatia_robust_2015,wan_robust_2017,dhar_modeling_2018,kuo_detecting_2018}.}}Many of these algorithms and recovery guarantees are based on techniques from compressed sensing (CS), and usually {{these works assume that the signal $\bfx$ is sparse over some basis, or generated from known physical models}}. In this paper, by applying techniques from deep learning, {{we will solve the outlier detection problem when $\bfx$ is generated from a generative model in deep learning}}. Namely, {{we}} propose an outlier detection method for signals modeled by neural network based generative models.

  {{Deep learning \cite{Goodfellow:2016} has attracted attentions}} from {{many}} fields in science and technology, and researchers have studied the signal recovery problem from the deep {{learning}} perspective, such as implementing traditional signal recovery algorithms by deep neural networks \cite{metzler_learned_2017,gupta_cnn-based_2018,yang_deep_2016}, and {{providing}} recovery guarantees for nonconvex optimization {{approaches}} in deep learning \cite{bora_compressed_2017,dhar_modeling_2018,wu_sparse_2018,papyan_theoretical_2018}. In \cite{bora_compressed_2017}, the authors {{considered}} the case where the measurements are contaminated with Gaussian noise {{of}} small magnitude, and they proposed to use a generative model to recover the signal from noisy measurements. {{Without}} requiring sparsity of the true signal, they showed that with high probability, the underlying signal can be recovered with small reconstruction error by $\ell_2$ minimization. However, similar to traditional CS, the techniques presented in \cite{bora_compressed_2017} can perform {{badly}} when the signal is corrupted with gross {{errors}}. Thus, it is of interest to study algorithms and theoretical performance guarantees of recovering signals modeled by deep generative models under outliers.

  In this paper, we propose an optimization framework for solving the outlier detection problem for signals generated from deep generative models. Instead of finding $\x\in\mathbb{R}^n$ directly from its compressed measurement $\y$ under the sparsity assumption of $\x$, we will find a signal $\z\in\mathbb{R}^k$ which can be mapped into $\x$ or a small neighborhood of $\x$ by a generator $G(\cdot)$. The generator $G(\cdot)$ is implemented by a neural network, and examples of generators {{include}} the variational auto-encoders or deep convolutional generative adversarial networks \cite{Goodfellow:2014,bora_compressed_2017,kingma_auto-encoding_2014}. Our results show that, for signals generated using a broad class of linear and non-linear generative models, $\ell_1$ minimization can perform outlier detection with theoretical performance guarantees. To the best of our knowledge, this is the first time that $\ell_1$ minimization is shown to provide performance guarantees for outlier detection for generative models, beyond the linear model considered in \cite{candes_decoding_2005}.

   Very often we can model signals by a generative model, which can be obtained by a training process using training data samples. In training, by using enough data samples $\{(\z^{(i)},\x^{(i)})\}_{i=1}^N$, we will obtain a prior a generator $G(\cdot)$ to map $\z^{(i)}\in\mathbb{R}^k$ to $G(\z^{(i)})$ such that $G(\z^{(i)})$ can approximate $\x^{(i)}\in\mathbb{R}^n$ well.  We can then use this known trained generator or generative model to perform outlier detection and find a low-dimensional $\z\in\mathbb{R}^k$ for previously unseen $\x\in\mathbb{R}^n$ such that $G(\z)$ approximates $\x$ well. 
   

  In this work, we first give necessary and sufficient conditions under which the true signal modeled by generative models can be recovered even under observations corrupted by outliers. We consider both the case where the generator is implemented by a linear neural network, and the case where the generator is implemented by a nonlinear neural network. In the {{linear neural network}} case, the generator is implemented by a $d$-layer neural network with random weight matrices $H_{i}$,where $i$ denotes the layer index, and identity activation functions in each layer. We show that, when the ground true weight matrices $H_{i}$ of the generator are random matrices and the generator has already been well-trained, {then we can theoretically guarantee successful recovery of the ground truth signal under sparse outliers via $\ell_0$ minimization with high probability}. Our results are also applicable to the nonlinear neural networks. Moreover, we give algorithms using $\ell_1$ minimization to recover the ground-truth signal modeled by generative models.  We further establish theoretical performance guarantees for the capability of $\ell_1$ minimization in performing outlier detection for generative models. In particular, we establish performance guarantees of $\ell_1$ minimization in performing outlier detection for  generative models employing leaky and regular ReLU activation functions. In some sense, the encoding matrix in \cite{candes_decoding_2005} can also be seen as a generator, but it is only a one-layer linear generator.  Our work can be seen as generalizing \cite{candes_decoding_2005} from linear generators to much more general non-linear multi-layer generators. 
  

  We summarize our contributions in this paper as follows: (1) We propose an outlier detection approach for signals modeled by generative models, {{ which further connects compressed sensing and deep learning; (2) We establish the recovery guarantees for the {{proposed approach}}, which hold for linear or nonlinear generative neural networks regardless of the depth of the neural networks; (3) We conduct extensive experiments to validate our theoretical analysis. We propose an iterative alternating direction method of multipliers algorithm for solving the outlier detection problem via $\ell_1$ minimization, and a gradient descent algorithm for solving a squared $\ell_1$ minimization formulation. 

  The rest of the paper is organized as follows. In Section \ref{Sec:ProblemStatement}, we give a formal statement of the outlier detection problem, and an overview of our outlier detection approach for signals modeled by generative models. In Section \ref{Sec:ADMMAlgorithm}, we propose the iterative alternating direction method of multipliers algorithm and the gradient descent algorithm. We present the recovery guarantees for generators implemented by both the linear and nonlinear neural networks in Section \ref{Sec:PerformanceAnalysis} using $\ell_0$ minimization. We provide analysis of recovery guarantees using $\ell_{1}$ minimization in Section \ref{Sec:leaky relu} and Section \ref{Sec:relu}. Experimental results are presented in Section \ref{Sec:NumericalResults}. We conclude this paper in Section \ref{Sec:ConclusionsandFutureDirections}.

  {{

  {\bf Notations:} {{In}} this paper, we will denote {{the number of nonzero elements of an vector $\x\in\mbR^n$ by $\|\x\|_0$}}, and the $\ell_1$ norm of an vector $\x\in\mbR^n$ by $\|\x\|_1 = \sum_{i=1}^n |\x_i|$. For an vector $\x\in\mbR^n$ and an index set $K$ with cardinality $|K|$, we {{let}} $(\x)_K\in\mbR^{|K|}$ {{denote}} the vector consisting of elements from $\x$ whose indices are in $K$. We let $K^c$ be the complement of set $K$ and {{we let}} $(\x)_{K^c}$ {{denote the}} vector with all the elements from $\x$ whose indices are not in $K$. {{We denote the signature of a permutation $\mu$ by ${\rm sign}(\mu)$}}. {{The value of ${\rm sign}(\mu)$ is $1$ or $-1$}}. The probability of an event $\mathcal{E}$ is denoted by $\mathbb{P}(\mathcal{E})$.

  }}


{\section{Problem Statement}\label{Sec:ProblemStatement}}

In this section, we will formally define the outlier detection problem for signals coming from generative models. 

Let generative model $G(\cdot):\mathbb{R}^k\to\mathbb{R}^n$ be implemented by a $d$-layer neural network. A conceptual diagram for illustrating the generator $G$ is given in Fig. \ref{fig:GeneralNNStructure}. We define the bias terms in each layer as $\boldsymbol{\delta_{1}}\in\mbR^{n_{1}}, \boldsymbol{\delta_{2}}\in\mbR^{n_{2}}, \cdots, \boldsymbol{\delta_{d-1}}\in\mbR^{n_{d-1}}, \boldsymbol{\delta_{d}}\in\mbR^{n}$, and the weight matrices $H_{i}$ for the $i$-th layer in \eqref{Defn:WeightMatrix}.
\begin{figure*}[!htb]
\begin{align}\label{Defn:WeightMatrix}
H_{1} = [H_{1(1)},H_{1(2)},\cdots,H_{1(n_{1})}]^T\in\mathbb{R}^{n_1 \times k}, H_{1(j)}\in\mbR^k, j=1,\cdots,n_1, \nonumber\\
H_{i}
= [H_{i(1)},H_{i(2)},\cdots,H_{i(n_{i})}]^T\in\mathbb{R}^{n_i \times n_{i-1}}, H_{i(j)}\in\mbR^{n_{i-1}}, j=1\cdots,n_i, 1< i < d,\nonumber\\
H_{d} = [H_{d(1)},H_{d(2)},\cdots,H_{d(n)}]^T\in\mathbb{R}^{n \times n_{d-1}}, H_{d(j)} \in\mbR^{n_{d-1}}, j=1,\cdots,n.
\end{align}
\end{figure*}

\begin{figure*}[!htb]
\begin{center}
\includegraphics[width=\textwidth, height=4.85cm]{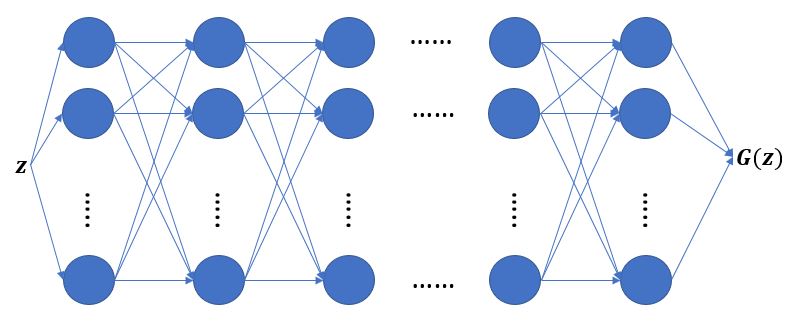}
\caption{General neural network structure of generator for mapping a low dimensional signal $\bfz$ to one in a higher dimensional space $G(\bfz)$.}
\label{fig:GeneralNNStructure}
\end{center}
\end{figure*}

The element-wise activation functions $\sigma(\cdot)$ in different layers are defined to be the same. Some commonly used activation functions include ReLU and leaky ReLU, i.e., 
\begin{align}
\sigma(\bfx_i) =
\begin{cases}
0, {\rm if\ } \bfx_i<0,\\
\bfx_i, {\rm if\ } \bfx_i\geq0
\end{cases}, 
\end{align}
and 
\begin{align}
\sigma(\bfx_i) =
\begin{cases}
\bfx_i, {\rm if\ } \bfx_i\geq 0,\\
h \bfx_i, {\rm if\ } \bfx_i<0,
\end{cases}
\end{align}
where $h\in(0,1]$ is a constant. Thus for a given input $\bfz\in\mathbb{R}^k$, the generative model's output $\x=G(\z)$ is given in \eqref{Defn:GeneratorOutput1}.

\begin{figure*}[!htb]
\begin{align}\label{Defn:GeneratorOutput1}
\x=G(\bfz) 
= \sigma(H_{d}\sigma(H_{d-1} \cdots \sigma(_{1}\bfz + \boldsymbol{\delta_{1}})  \cdots + \boldsymbol{\delta_{d-1}})+ \boldsymbol{\delta_{d}}).
\end{align}
\end{figure*}


The measurement vector $\bfy\in\mbR^{m}$ is given as $\bfy = \mcM\bfx + \bfe$, 
where the $\mcM\in\mbR^{m\times n}$ is a measurement matrix, {$\bfx$} is the signal to be recovered, and the $\bfe$ is the vector of outliers occurring due to the corruption of measurements.  We assume that $\bfe$ is a sparse vector with many zero elements since the number of outliers is often small. 

The signal $\bfx$ can be recovered via the following $\ell_0$ ``norm'' minimization if $\x$ is modeled by the generator $G(\cdot)$:
\begin{align}\label{Defn:L0Minimization}
\min_{\bfz\in\mbR^k} \|\mcM G(\bfz) - \bfy\|_0.
\end{align}
However, the aforementioned $\ell_0$ ``norm'' minimization is an NP-hard problem, thus we relax it to the following $\ell_1$ minimization for recovering $\x$, i.e.,
\begin{align}\label{Defn:L1Minimization}
      \min_{\z\in\mathbb{R}^k} \|{\mcM}G(\z) - \y\|_1.
\end{align}
When the generator is well-trained, it can map low-dimension signal $\z\in\mathbb{R}^k$ to high-dimension signal $G(\z)\in\mathbb{R}^n$. These well-trained generators will be used to solve the outlier detection problem.

Our goal is to find a $\z\in\bfR^{k}$ such that $\hat{\x} = G(\z) \in\bfR^{n}$ satisfying the following property: when $\hat{\x}$ goes through the same measurement process, the result $\hat{\y}\in\bfR^{m}$ will be close to the measurement $\y$ resulting from the ground-truth signal $\x$.  


Suppose that $G(\cdot)$ is a linear mapping, this problem can be reduced to the linear decoding problem considered in \cite{candes_decoding_2005}. In this case, \eqref{Defn:L1Minimization} can be solved by linear programming, as shown in \cite{candes_decoding_2005}. However, in our problem, $G(\cdot)$ is a non-linear mapping, and we need to design new numerical optimization algorithms to solve \eqref{Defn:L1Minimization}.

Motivated by linearization techniques, we propose to solve the above $\ell_1$ minimization by alternating direction method of multipliers (ADMM) algorithm which is introduced in Section \ref{Sec:ADMMAlgorithm}. We also propose to solve the outlier detection problem by the gradient descent algorithm for minimizing the squared $\ell_1$ minimization, i.e.,
      \begin{align}\label{Defn:SquaredL1Minimization}
      \min_{\bfz\in\mbR^k} \|\mcM G(\bfz) - \bfy\|^2_1.
      \end{align}
We note that \eqref{Defn:L1Minimization} and \eqref{Defn:SquaredL1Minimization} are equivalent optimization problems.

{{
{\bf Remarks:} In analysis, we focus on recovering signal and detecting outliers. However, our experimental results consider the signal recovery problem with the presence of both outliers and noise.
}}



Note that in \cite{dhar_modeling_2018}, the authors {considered} the following problem
  \begin{align*}
  \min_{\bfz,\bfe} \|\bfe\|_0, {\rm s.t.\ }\mcM(G(\bfz)+\bfe) = \bfy.
  \end{align*}
  Our problem is fundamentally different from the above problem. On the one hand, although the problem in \cite{dhar_modeling_2018} can also be treated as an outlier detection problem, the outlier occurs in {{signal $\bfx$ itself}}. In contrast, in our problem, the {{outliers}} appear in the measurement process. On the other hand, {{our analytical recovery performance guarantees are very different, and so are our analytical techniques.
  }}

 In our numerical results, we also compare the performance of \eqref{Defn:L1Minimization} against the traditional $\ell_2$ minimization and regularized $\ell_2$ minimization described as follows: 
\begin{align}\label{Defn:SquaredL2Minimization}
\min_{\z\in\mathbb{R}^k} \|{\mcM}G(\z) - \y\|_2^2,
\end{align}
and
\begin{align}\label{Defn:RegularizedSquaredL2Minimization}
\min_{\z\in\mathbb{R}^k} \|{\mcM}G(\z) - \y\|_2^2 + \lambda \|\z\|_2^2.
\end{align}
Both the above two $\ell_{2}$ norm minimization problems can be solved by gradient descent solvers \cite{Barzilai88}.

{\section{Solving $\ell_1$ Minimization via Gradient Descent and Alternating Direction Method of Multipliers}\label{Sec:ADMMAlgorithm}}

In this section, we introduce both gradient descent algorithms and alternating direction method of multipliers (ADMM) algorithms to solve the outlier detection problem. We consider the $\ell_1$ minimization (\ref{Defn:L1Minimization}) in Section \ref{Sec:ProblemStatement}.  Namely, for $\ell_1$ minimization, we consider both the case where we solve (\ref{Defn:L1Minimization}) by the ADMM algorithm introduced, and the case where we solve (\ref{Defn:SquaredL1Minimization}) by the gradient descent (GD) algorithm \cite{Barzilai88}.

\subsection{Gradient descent for squared $\ell_1$ minimization}
Theoretically speaking, due to the non-differentiability of the absolute value function at $0$, the gradient descent method cannot be applied to solve the problem (\ref{Defn:L1Minimization}). However, we observe that if we turn to solve the equivalent squared $\ell_1$ norm minimization (\ref{Defn:SquaredL1Minimization}), the gradient descent algorithm can work well in practice though the squared $\ell_1$ norm is not differentiable at a set of measure $0$. Please see Section \ref{Sec:NumericalResults} for experimental results.

\subsection{ADMM algorithm for solving \eqref{Defn:L1Minimization}}
Motivated by linearization technique and the ADMM technique for solving basis pursuit problem in \cite{candes_decoding_2005}, we propose an iterative linearized ADMM algorithm for solving (\ref{Defn:L1Minimization}) where the nonlinear mapping $G(\cdot)$ at each iteration is locally approximated via a linear mapping. Specifically, we introduce an auxiliary variable $\bfw$, and the above problem can be re-written as
\begin{align}
\min_{\textbf{z}\in\mathbb{R}^k,\textbf{w}\in\mathbb{R}^{m}} \|\textbf{w}\|_1\nonumber, 
{\rm s.t.\ }{\mcM}G(\textbf{z})-\textbf{w}=\textbf{y}.
\end{align}

The augmented Lagrangian of the above problem is given in \eqref{Eq:AugmentedLagrangian}, 
\begin{figure*}[!htb]
\begin{align}\label{Eq:AugmentedLagrangian}
L_{\rho}(\bfz,\bfw,\lambda)
& = \|\bfw\|_{1} + \lambda^{T}(\mcM G(\bfz) - \bfw - \bfy)
+ \frac{\rho}{2} \| \mcM G(\bfz) - \bfw - \bfy\|_{2}^{2} \nonumber\\
& = \|\bfw\|_1
+ \frac{\rho}{2}\left\|\mcM G(\bfz) - \bfw - \bfy + \frac{{\bf \lambda}}{\rho}\right\|^2_2 
-\frac{\rho}{2} \|\lambda/\rho\|_{2}^{2},
\end{align}
\end{figure*}
where {{${\bf \lambda}\in\mathbb{R}^m$}} and $\rho\in\mathbb{R}$ are Lagrangian multipliers and penalty parameter respectively. The main idea of ADMM algorithm is to {update $\bfz$ and $\bfw$ alternatively}. Notice that for both $\bfw$ and ${\bf \lambda}$, they can be updated following the standard procedures. The nonlinearity and the lack of explicit form of the generator make it nontrivial to find the updating rule for $\bfz$. Thus, we consider a local linearization technique to solve the problem.

Let us consider the $q$-th iteration for $\bfz$, namely we are trying to find
${{\bf z}^{q + 1}}
= \arg \mathop {\min }\limits_{\bf z} {L_\rho }\left( {{\bf z},{{\bf w}^q},{\lambda ^q}} \right) 
= \arg \mathop {\min }\limits_{\bf z} F(\bfz)$, 
where $F\left( {\bf z} \right)$ is defined as $F(\bfz) = \left\| {{{{\mcM}}}G\left( {\bf z} \right) - {{\bf w}^q} - {\bf y} {{+}} \frac{{{\lambda ^q}}}{\rho }} \right\|_2^2$. Since $G({\bf z})$ is non-linear, we will use a first order approximation, i.e.,
$G\left( {\bf z} \right)\approx G({\bf z}^q)+\nabla G({\bf z}^q)^T({\bf z}-{\bf z}^q)$,
where $\nabla G({\bf z}^q)^T$ is the transpose of the gradient of $G(\cdot)$ at ${\bf z}^q$. Then $F({\bf z})$ can be formulated as in \eqref{Eq:FzReformulation}.

\begin{figure*}[!htb]
\begin{align}\label{Eq:FzReformulation}
{
F\left( {\bf z} \right)}
& {=\left\| {{\mcM}\left( G({\bf z}^q)+\nabla G({\bf z}^q)^T ({\bf z}-{\bf z}^q) \right) - {{\bf w}^q} - {\bf y} {{+}} \frac{{{\lambda ^q}}}{\rho }} \right\|_2^2} \nonumber\\
& {=\left\| {\mcM}\nabla G({\bf z}^q)^T{\bf z}+{\mcM}\left( G({\bf z}^q)-\nabla G({\bf z}^q)^T {\bf z}^q\right) - { {{\bf w}^q}} - {\bf y} {{+}} \frac{{{\lambda ^q}}}{\rho } \right\|_2^2.}
\end{align}
\end{figure*}

The considered problem is a least square problem, and {the minimum is achieved at \eqref{Eq:Minimizer},
\begin{figure*}[!htb]
\begin{align}\label{Eq:Minimizer}
\bfz^{q+1}
 =
(\mcM\nabla G({\bf z}^q)^T )^{\dagger} 
\times\left({\bf w}^q +{\bf y} - \frac{\lambda ^q}{\rho}-\left(\mcM G({\bf z}^q)-\mcM\nabla G({\bf z}^q)^T {\bf z}^q\right)\right)
\end{align}
\end{figure*}
where the $\dagger$ denotes the pseudo-inverse.} Notice that $G(z^q)$ can be obtained from the output of generator, and the gradient of $G(z^q)$ can be calculated by compute\_gradient function from Tensorflow library \cite{abadi_tensorflow:_2016}. Similarly, we can update $\bfw$ and ${\bf \lambda}$ by
\begin{align*}
{{\bf w}^{q + 1}} 
& = \arg \mathop {\min }\limits_{\bf w} \frac{\rho }{2}\left\| {{{{\mcM}}}G\left( {\bfz^{q+1}}\right) - {{\bf w}} - {\bf y} {{+}} \frac{{{\lambda ^q}}}{\rho }} \right\|_2^2 + {\left\| {\bf w}\right\|_1} \\
& = {T_{\frac{1}{\rho }}}\left( {{\mcM}G\left( {{{\bf z}^{q + 1}}} \right) - {\bf y} {{+}} \frac{{{\lambda ^q}}}{\rho }} \right),
\end{align*}
and
\begin{align*}
{\lambda ^{q + 1}} = {\lambda ^q} + \rho \left( {{\mcM}G\left( {{{\bf z}^{q + 1}}} \right) - {{\bf w}^{q + 1}} - {\bf y}} \right),
\end{align*}
where ${T_{\frac{1}{\rho }}}$ is the element-wise soft-thresholding operator with parameter $\frac{1}{\rho}$ \cite{Boyd:2011}, i.e.,
\begin{align*}
[T_{\frac{1}{\rho}}(\bfx)]_i
=
\begin{cases}
[\bfx]_i - \frac{1}{\rho}, [\bfx]_i> \frac{1}{\rho}, \\
0, |[\bfx]_i| \leq \frac{1}{\rho}, \\
[\bfx]_i + \frac{1}{\rho}, [\bfx]_i < - \frac{1}{\rho}.
\end{cases}
\end{align*}
}}

We continue the iterative procedures to update $\bf z$ until the optimization problem {{converges}} or stopping criteria are met. More discussions on stopping criteria and parameter tuning can be found in \cite{Boyd:2011}. {{Our algorithm is summarized in Algorithm \ref{Alg:IADMM}.}} The final value ${\bf z}^*$ can be used to generate the estimate $\hat {\bf x} = G({\bf z^*})$ for the true signal. We define the following metrics for evaluation: the measured error caused by imperfectness of CS measurement, i.e., ${\varepsilon _m} \triangleq \left\| {\textbf{y} - {\mcM}G\left( {{\bf{\hat z}}} \right)} \right\|_1$, and reconstructed error via a mismatch of $G(\hat{\bf z})$ and $\x$, i.e., ${\varepsilon _r} \triangleq \left\| {{\bf{x}} - G\left( {{\bf{\hat z}}} \right)} \right\|_2^2.$

\begin{algorithm}
\caption{{Linearized ADMM}}\label{Alg:IADMM}
\begin{algorithmic}

\STATE {\bf Input:} $\bfz^0, \bfw^0, {\bf \lambda}^0, \mcM$

\STATE {\bf Parameters:} $\rho$, $MaxIte$

\STATE $q \gets 0$

\WHILE{$q\leq MaxIte$}

\STATE $A^{q+1} \gets \mcM \nabla G(\bfz^{q})^T$

\STATE $\bfz^{q+1} \gets {(A^{q+1})^{\dagger}}(\bfw^q + \bfy - {\bf\lambda}^q/\rho - (\mcM G(\bfz^q) - A^{q+1} \bfz^q))$

\STATE $\bfw^{q+1} \gets T_{\frac{1}{\rho}}\left( \mcM G(\bfz^{q+1}) - \bfy + \frac{{\bf \lambda}^q}{\rho} \right)$

\STATE ${\bf\lambda}^{q+1} \gets {\bf\lambda}^q + \rho (\mcM G(\bfz^{q+1}) - \bfw^{q+1} - \bfy)$

\ENDWHILE

\STATE {\bf Output:} $\z^{MaxIte+1}$

\end{algorithmic}
\end{algorithm}

\section{Performance Analysis: $\ell_0$ minimization}\label{Sec:PerformanceAnalysis}

In this section, we present our theoretical analysis {{of}} the performance of outlier detection for signals modeled by generative models. We first establish the necessary and sufficient condition for recovery in {{Theorems}} \ref{Thm:L0MinimizationRecoveryCondition} for $\ell_0$ minimization. Note that this condition appeared in \cite{xu_sparse_2013}, and we present this condition and its proof here for the sake of completeness. We then show that both the linear and nonlinear neural network with {{an}} arbitrary {{number of}} layers can satisfy the recovery condition, thus {ensuring} successful reconstruction of signals.

\subsection{Preliminaries}

Let the ground truth signal $\textbf{x}_0\in\mathbb{R}^n$, the measurement matrix ${\mcM}\in\mathbb{R}^{m\times n} ({m < n})$, and the sparse outlier vector $\textbf{e}\in\mathbb{R}^{{m}}$ be such that $\|\textbf{e}\|_0\leq {l < m}$, then the problem can be stated as
\begin{align}\label{Defn:L0MinimizationNew}
\min_{\textbf{z}\in\mathbb{R}^k} \|{\mcM}G(\textbf{z}) -\textbf{ y}\|_0
\end{align}
where $\textbf{y} = {\mcM}\textbf{x}_0+\textbf{e}$ is a measurement vector and $G(\cdot):\mathbb{R}^k\rightarrow \mathbb{R}^n$ is a generative model. Let $\textbf{z}_0\in\mathbb{R}^k$ be a vector such that $G(\textbf{z}_0)=\textbf{x}_0$. Then we have following conditions under which $\textbf{z}_0$ can be recovered exactly without any sparsity assumption on neither $\textbf{z}_0$ nor $\textbf{x}_0$.

\begin{theorem}\label{Thm:L0MinimizationRecoveryCondition}
  Let $G(\cdot), \textbf{x}_0, \textbf{z}_0, \textbf{e}$ and $\textbf{y}$ be defined as above. The vector $\textbf{z}_0$ can be recovered exactly from (\ref{Defn:L0MinimizationNew}) {for any $\textbf{e}$ with $\|\textbf{e}\|_0\leq l$} if and only if $\|{\mcM}G(\textbf{z})-{\mcM}G(\textbf{z}_0)\|_0 \geq 2l+1$ holds for any $\textbf{z}\neq \textbf{z}_0$.
\end{theorem}

\begin{proof}
    We first show the sufficiency. Assume $\|{\mcM}G(\textbf{z})-{\mcM}G(\textbf{z}_0)\|_0 \geq 2l+1$ holds for any $\textbf{z}\neq \textbf{z}_0$. The triangle inequality gives
  \begin{align*}
  \|{ \mcM}G(\textbf{z}) - {\mcM}G(\textbf{z}_0)\|_0
  & \leq \|{ \mcM}G(\textbf{z}_0)-\textbf{y}\|_0 \\
  &\quad + \|{ \mcM}G(\textbf{z})-\textbf{y}\|_0.
  \end{align*}
  So we have
  \begin{align*}
  \|{ \mcM}G(\textbf{z}) - \textbf{y}\|_0
  & \geq \|{ \mcM}G(\textbf{z}) - { \mcM}G(\textbf{z}_0)\|_0 \\
  &\quad - \|{ \mcM}G(\textbf{z}_0)-\textbf{y}\|_0 \\
  & \geq (2l+1)-l 
  = l+1  \\
  &  > \|{ \mcM}G(\textbf{z}_0) - \textbf{y}\|_0.
  \end{align*}
  This means $\textbf{z}_0$ is {{a}} unique optimal solution to (\ref{Defn:L0MinimizationNew}).
  
  Next, we show the necessity by contradiction. Assume $\|{ \mcM}G(\textbf{z}) - { \mcM}G(\textbf{z}_0)\|_0\leq 2l$ holds for certain $\textbf{z}\neq \bfz_0\in\mathbb{R}^k$, and this means ${ \mcM}G(\textbf{z})$ and ${ \mcM}G(\textbf{z}_0)$ differ from each other over at most $2l$ entries. Denote by $\mathcal{I}$ the index set where ${ \mcM}G(\textbf{z}) - { \mcM}G(\textbf{z}_0)$ has nonzero entries, then $|\mathcal{I}|\leq 2l$. We can choose outlier vector $\textbf{e}$ such that $\textbf{e}_i=({ \mcM}G(\textbf{z}) - { \mcM}G(\textbf{z}_0))_i$ for all $i\in\mathcal{I}'\subset \mathcal{I}$ with $|\mathcal{I}'|= l$, and $\textbf{e}_i=0$ otherwise. Then
  \begin{align*}
  \|{ \mcM}G(\textbf{z}) - \textbf{y}\|_0
  & = \|{ \mcM}G(\textbf{z}) - { \mcM}G(\textbf{z}_0) - \bfe\|_0 \\
  & \leq 2l - l 
  =l = \|{ \mcM}G(\textbf{z}_0) - \textbf{y}\|_0.
  \end{align*}
  This means that $\bfz_0$ is not a unique solution to (\ref{Defn:L0MinimizationNew}).
\end{proof}




Before we proceed to the main results, we give some technical lemmas which will be used to prove our main theorems.

\begin{lemma}\label{Lem:Schwartz-ZippelLemma}
(\cite{zippel_effective_2012}) Let $P\in A[X_1,\cdots,X_v]$ be a polynomial of total degree $D$ over an integral domain $A$. Let $\mathscr{J}$ be a subset of $A$ of cardinality $B$. Then
$\mathbb{P}(P(x_1,\cdots,x_v)=0|x_i\in\mathscr{J}) \leq \frac{D}{B}.$
\end{lemma}

Lemma \ref{Lem:Schwartz-ZippelLemma} gives an upper bound of the probability for a polynomial to be zero when its variables are randomly taken from a set. Note that in \cite{khajehnejad_sparse_2011}, the authors applied Lemma \ref{Lem:Schwartz-ZippelLemma} to study the adjacency matrix from an expander graph. Since the determinant of a square random matrix is a polynomial, Lemma \ref{Lem:Schwartz-ZippelLemma} can be applied to determine whether a square random matrix is rank deficient. An immediate result will be Lemma \ref{Lem:FullRankOfGaussianRandomMatrix}.

\begin{lemma}\label{Lem:FullRankOfGaussianRandomMatrix}

A random matrix $M\in\mathbb{R}^{m\times n}$ with independent Gaussian entries will have full rank with probability $1$, i.e., ${\rm rank}(M)=\min(m,n)$.

\end{lemma}

\begin{proof}
  For simplicity of presentation, we consider the square matrix case, i.e., a random matrix $M\in\mathbb{R}^{n\times n}$ whose entries are drawn i.i.d. randomly from $\mathbb{R}$ according to the  standard Gaussian distribution $\mathcal{N}(0,1)$. Note that
  $$
  {\rm det}(M) = \sum_{\mu} \left({\rm sign}(\mu) \prod_{i=1}^n M_{i\mu(i)}\right)
  $$
  where $\mu$ is a permutation of $\{1,2,\cdots,n\}$, the summation is over all $n!$ permutations, and ${\rm sign}(\mu)$ is either $+1$ or $-1$. Since ${\rm det}(M)$ is a polynomial of degree $n$ with respect to $M_{ij}\in\mathbb{R}$, $i,j=1,\cdots,n$ and $\mathbb{R}$ {{has infinitely many elements}}, then according to Lemma \ref{Lem:Schwartz-ZippelLemma}, we have $\mathbb{P}({\rm det}(M)=0) \leq 0$. Thus, 
  $\mathbb{P}(M\text{ has full rank})
  = \mathbb{P}({\rm det}(M)\neq 0)
  \geq 1 - \mathbb{P}({\rm det}(M)=0),
  $
  and this means the Gaussian random matrix has full rank with probability $1$.
\end{proof}

{\bf Remarks:} (1) We can easily extend the arguments to random matrix with arbitrary shape, i.e., $M\in\mathbb{R}^{m\times n}$. {{Consider}} arbitrary square sub-matrix with size $\min(m,n)\times \min(m,n)$ from $M$, following the same arguments will give that with probability 1, the matrix $M$ has full rank with ${\rm rank}(M)=\min(m,n)$; (2) The arguments in Lemma \ref{Lem:FullRankOfGaussianRandomMatrix} can also be {{extended}} to random matrix with other distributions.

\subsection{Generative {{Models}} via Linear Neural {{Networks}}}

We first consider the case where the activation function is an element-wise identity operator, i.e., $\sigma(\bfx)=\bfx$, and this will give us a simplified input and output relation as follows
\begin{align}\label{Eq:LinearNN}
G(\bfz) = H_{d}H_{d-1}\cdots H_{1}\bfz.
\end{align}
Define $W = H_{d}H_{d-1}\cdots H_{1}$, we get a simplified relation $G(\bfz)=W\bfz$. We will show that with high probability: for any $\bfz\neq \bfz_0$, the $\ell_0$ norm 
$\|G(\bfz) - G(\bfz_0)\|_0
= \|W(\bfz - \bfz_0)\|_0$
will have at least $2l+1$ nonzero entries; or when we treat the $\z-\z_0$ as a whole and still use the letter $\z$ to denote $\z-\z_0$, then for any $\bfz\neq 0$, the vector $\bfv := W\bfz$ will have at least $2l+1$ nonzero entries.

Note that in the above derivation, the bias term is incorporated in the weight matrix. For example, in the first layer, we know
$
H_{1}\bfz + \boldsymbol{\delta_{1}}
= [H_{1}\ \boldsymbol{\delta_{1}}]
\left[\begin{matrix}
\bfz^T & 1
\end{matrix}\right]^T.
$
Actually, when the $H_{i}$ does not incorporate the bias term $\boldsymbol{\delta_{i}}$, we have a generator as in \eqref{Eq:GzReformulation}.
\begin{figure*}
\begin{align}\label{Eq:GzReformulation}
 G(z)
&  = H_{d}(H_{d-1}(
\cdots (H_{2}(H_{1}{\bf z} + \boldsymbol{\delta_{1}}) + \boldsymbol{\delta_{2}})+\cdots) + \boldsymbol{\delta_{d-1}}) + \boldsymbol{\delta_{d}}   \nonumber \\
& = H_{d}\cdots H_{1}\z
+ H_{d}\cdots H_{2} \boldsymbol{\delta_{1}}
+ H_{d}\cdots H_{3} \boldsymbol{\delta_{2}} + \cdots + \boldsymbol{\delta_{d}}.
\end{align}
\end{figure*}
Then in this case, we need to show that with high probability: for any $\z\neq \z_0$, the $\ell_0$ norm
$\|G(\z) - G(\z_0)\|_0 = \|W(\z - \z_0)\|_0$
{has at least $2l+1$ nonzero entries because all the other terms are cancelled.}

{{
\begin{lemma}\label{Lem:FullRankAndZeroEntries}
Let matrix $W\in\mbR^{n\times k}$. Then for an integer $l \geq 0$ , if every $r:= n-(2l+1)\geq k$ rows of $W$ has full rank $k$, then $\bfv =W \bfz$ can have at most $n-(2l+1)$ zero entries {{for every}} $\bfz\neq 0\in\mathbb{R}^k$.
\end{lemma}
}}

\begin{proof}
  Suppose $\bfv = W\bfz$ has $n-(2l+1)+1$ zeros. {{We}} denote by $S_\bfv$ the set of indices where the corresponding entries of $\bfv$ are nonzero, and by $\bar{S}_\bfv$ the set of indices where the corresponding entries of $\bfv$ are zero. Then $|\bar{S}_\bfv|=n-(2l+1)+1$. We thus have the linear equations
  $${
    \bm{0} = \bfv_{\bar{S}_\bfv} = W_{\bar{S}_\bfv} \bfz,
  }$$
  where $W_{\bar{S}_\bfv}$ consists of $|\bar{S}_\bfv| = n-(2l+1) + 1$ rows from $W$ with indices in $\bar{S}_\bfv$. Then the rank of the matrix $W_{\bar{S}_\bfv}$ be smaller than $k$ so that there exists {{such}} a nonzero solution $\bfz$. This means there {{exists}} a sub-matrix consisting of $n-(2l+1)$ rows which has rank smaller than $k$, contradicting the rank assumption in this lemma.
\end{proof}

Lemma \ref{Lem:FullRankAndZeroEntries} actually gives a sufficient condition for $\bfv = W\bfz$ to have at least $2l+1$ nonzero entries, i.e., every $n_d-(2l+1)$ rows of $W$ has full rank $k$.

\begin{lemma}\label{Lem:LinearNNArbitraryLayers}

Let the composite weight matrix in multiple-layer neural network with identity activation function be $W = H_{d}H_{d-1}\cdots H_{1}$, 
where $H_{i}, i=1,\cdots,d$ are defined as in (\ref{Defn:WeightMatrix}).  Assume $n_i\geq k, i=1,\cdots,d-1$  and {{$n> k$}}. When $d=1$, we let $W = H_{1}\in\mathbb{R}^{n\times k}$. Let each entry in each weight matrix $H_{i}$ be drawn independently randomly according to {{the standard Gaussian distribution, $\mcN(0,1)$}}. Then for a linear neural network {{of}} two or more layers {{with}} composite weight matrix $W$ defined above, with probability 1, every $r=n_d-(2l+1)\geq k$ rows of $W$ will have full rank $k$.

\end{lemma}

\begin{proof}
  
  We will show Lemma \ref{Lem:LinearNNArbitraryLayers} by induction.

{\bf $1$-layer case:} By Lemma \ref{Lem:FullRankOfGaussianRandomMatrix}, with probability $1$, the matrix $W=H_{1}$ has full rank $k$, where $H_{1}\in\mathbb{R}^{n_1\times k}$ ($n_1\geq k$). Now take $r$ rows of $H_1$ to form a new matrix $M^{(1)}\in \mathbb{R}^{r\times k}$, then $M^{(1)}$ still has full rank $k$ with probability $1$.

{\bf $(d-1)$-layer case:} In this case, $W=H_{d-1}H_{d-2}\dotsb H_1$. Assume that the matrix $W$ has full rank $k$ with probability 1, and every $r$ rows of $W$ also has full rank $k$ with probability 1.

{\bf $d$-layer case:} $H_{d-1}H_{d-2}\dotsb H_1$ has singular value decomposition $H_{d-1}H_{d-2}\dotsb H_1=U \Sigma V^*$, where $\Sigma\in\mathbb{R}^{k\times k}$ and $V\in\mathbb{R}^{k\times k}$ have rank $k$, and $U\in\mathbb{R}^{n_{d-1}\times k}$.
We take arbitrary $r$ rows of matrix $H_{d}$ to form a new matrix $M^{(d)}$.
Then $M^{(d)}H_{d-1}H_{d-2}\dotsb H_1=M^{(d)}U \Sigma V^*$. Since $\Sigma$ and $V$ have full rank $k$, then ${\rm rank}(M^{(d)}H_{d-1}H_{d-2}\dotsb H_1)
  = {\rm rank}(M^{(d)}U)$. Since each element of $M^{(d)}$ is i.i.d. Gaussian, and the columns of $U$ are orthonormal columns, each element of $M^{(d)}U$ is still i.i.d. Gaussian. By Lemma \ref{Lem:FullRankOfGaussianRandomMatrix}, with probability $1$, $M^{(d)}U$ has full rank $k$.

  
  Define the following event as
  $\mathcal{E}$: there is a matrix $M^{(d)}$ such that $M^{(d)}H_{d-1}H_{d-2}\dotsb H_{1}$ is rank deficient.
  
  Notice that we have totally ${n_d\choose{n_d-(2l+1)}}$ choices for forming $M^{(d)}$, thus according to the union bound, the probability for existence of a matrix $M^{(d)}$ which makes $M^{(d)}H_{d-1}H_{d-2}\dotsb H_{1}$ rank deficient satisfies
      \begin{align*}
      \mbP(\mathcal{E})
      \leq {n_d\choose{n_d-(2l+1)}} \times 0
      =0.
      \end{align*}
      Thus, the probability for all such matrices $M^{(d)}H_{d-1}H_{d-2}\dotsb H_{1}$ to have full rank is $1$.

\end{proof}


Thus, we can conclude that for every $\bfz\neq 0\in\mathbb{R}^k$, if the weight matrices are defined {{as}} above, then the $\y=W \bfz\in\mathbb{R}^n$ will have at least $2l+1$ nonzero elements, and the successful recovery is guaranteed by the following theorem.

\begin{theorem}\label{Thm:LinearNNRecoveryGuarantee}

Let the generator $G(\cdot)$ be implemented by a multiple-layer neural network with identity activation function. The weight matrix in each layer is defined as in (\ref{Defn:WeightMatrix}). Assume $n_i\geq k, i=1,\cdots,d-1$  and {{$n=n_d> k$}}. Each entry in each weight matrix $H_{i}$ is independently randomly drawn from standard Gaussian distribution $\mcN(0,1)$. Let the measurement matrix $M\in\mathbb{R}^{m\times n}$ be a random matrix with each entry i.i.d. drawn according to standard Gaussian distribution $\mcN(0,1)$. Let the true signal $\bfx_0$ and its corresponding low dimensional signal $\bfz_0$, the outlier signal $\bfe$ and $\y$ be the same as in Theorem \ref{Thm:L0MinimizationRecoveryCondition} and {{satisfy
$
\|\bfe\|_{0} \leq (m-1-k)/2.
$
}}
Then with probability $1$, every possible $\bfz_0$
can be recovered from (\ref{Defn:L0MinimizationNew}).

\end{theorem}

\begin{proof}
  Notice that in this case, we have
  $MG(\bfz) = MH_{d}H_{d-1}\cdots H_{1}\bfz.$
  Since $M^{m\times n}$ is also a random matrix with i.i.d. Gaussian entries, we can actually treat $MG(\bfz)$ as a $(d+1)$-layer neural network implementation of the generator acting on $\bfz$. Let $l\leq \frac{m-1-k}{2}$, then {{according}} to Lemma \ref{Lem:LinearNNArbitraryLayers}, every ${m}-(2l+1)$ rows of $MH_{d}H_{d-1}\cdots H_{1}$ will have full rank $k$ with probability $1$. Lemma \ref{Lem:FullRankAndZeroEntries} implies that the $MH_{d}H_{d-1}\cdots H_{1}(\z-\z_0)$ can have at most ${m}-(2l+1)$ zero entries for every $\z\neq \z_0$. Finally, from Theorem \ref{Thm:L0MinimizationRecoveryCondition}, $\bfz_0$ can be recovered with probability 1.
\end{proof}

\subsection{Generative {{Models}} {{via}} Nonlinear Neural {{Network}}}

We have shown that when the activation function $\sigma(\cdot)$ is the identity function, with high probability the generative model implemented via neural network can successfully recover the true signal $\bfz_0$. In this section, we will extend our analysis to neural networks with nonlinear activation functions.

{{
  Notice that in the proof of neural {{networks}} with identity activation, the key point is that we can write
  $G(\bfz) - G(\bfz_0)$ as $W(\bfz-\bfz_0)$, which allows us to characterize the recovery conditions by the properties of linear equation systems. However, in neural {{network}} with nonlinear activation functions, we do not have such linear properties. For example, in the case {{of}} leaky ReLU activation function, the sign patterns of $\bfz$ and $\bfz_0$ can be different. Thus we cannot directly transform $G(\bfz) - G(\bfz_0)$ into $W(\bfz - \bfz_0)$. Fortunately, we can achieve the transformation via Lemma \ref{Lem:PatternIrrelevanceInLeakyReLU} for the leaky ReLU activation function.

\begin{lemma}\label{Lem:PatternIrrelevanceInLeakyReLU}
 Denote $\sigma$ as leaky ReLU activation function which was defined previously, then for arbitrary $x\in\mbR$ and $y\in\mbR$, the $\sigma(x)-\sigma(y)=\beta(x-y)$ holds for some {{$\beta\in[h,1]$}} regardless of the sign patterns of $x$ and $y$.
\end{lemma}

\begin{proof}
  When $x$ and $y$ have the same sign pattern, i.e., $x\geq 0$ and $y\geq 0$ holds simultaneously, or $x<0$ and $y<0$ holds simultaneously, then $\sigma(x) - \sigma(y) = x - y$ with $\beta=1$, or $\sigma(x) - \sigma(y) = h(x-y)$ with $\beta=h\in(0,1]$.
  
  When $x$ and $y$ have different sign patterns, we have the following two cases: (1) $x\geq 0$ and $y<0$ holds simultaneously; (2) $x<0$ and $y\geq 0$ holds simultaneously. For the first case, we have
  $\sigma(x) - \sigma(y)
  = x - hy
  = \beta (x-y)$, {{which gives
      $
      \beta = \frac{x-hy}{x-y}
      \geq \frac{hx - hy}{x-y}
      = h,
      $
      and
      $
      \beta = \frac{x-hy}{x-y}
      \leq \frac{x-y}{x-y} = 1,
      $
      where we use the facts that  $x\geq 0$, $y<0$ and $h\in(0,1]$. Thus $\beta\in[h,1]$. Similar arguments work for $x<0$ and $y\geq 0$. 
      
      Combine the above cases together, we have $\beta\in[h,1]$.}}
\end{proof}

Since in the neural network, the leaky ReLU acts on the weighted sum $H\bfz$ {{element-wise}}, we can easily extend Lemma \ref{Lem:PatternIrrelevanceInLeakyReLU} to  the high dimensional case. For example, consider the first layer with weight $H_{1}\in\mbR^{{n_1}\times k}$, bias $\boldsymbol{\delta_{1}}\in\mbR^{{n_1}}$, and leaky ReLU activation function, {{we have
\begin{align*}
& \sigma_{1}(H_{1}\bfz + \boldsymbol{\delta_{1}}) - \sigma_{1}(H_{1}\bfz_0 +\boldsymbol{\delta_{1}} ) \\
& =
\left[\begin{matrix}
\beta^{(1)}_1 ([H_{1}\bfz + \boldsymbol{\delta_{1}}]_1 - [H_{1}\bfz_0 + \boldsymbol{\delta_{1}}]_1) \\
\beta^{(1)}_2 ([H_{1}\bfz + \boldsymbol{\delta_{1}}]_2 - [H_{1}\bfz_0 + \boldsymbol{\delta_{1}}]_2) \\
\vdots\\
\beta^{(1)}_{{n_1}} ([H_{1}\bfz + \boldsymbol{\delta_{1}}]_{{n_1}} - [H_{1}\bfz_0 + \boldsymbol{\delta_{1}}]_{{n_1}})
\end{matrix}\right] \\
& =
\left[\begin{matrix}
\beta^{(1)}_1 & 0 &\cdots &0\\
0 & \beta^{(1)}_2 &\cdots &0\\
\vdots &\vdots &\ddots &0\\
0 & 0 & \cdots &\beta^{(1)}_{{n_1}}
\end{matrix}\right] \\
&\quad \times 
(H_{1}\bfz + \boldsymbol{\delta_{1}} - (H_{1}\bfz_0 + \boldsymbol{\delta_{1}})) \\
& = \Gamma^{(1)} H_{1}(\bfz-\bfz_0),
\end{align*}
where $\beta^{(1)}_i\in[h,1]$ for every $i=1,\cdots,{n_1}$, and $\Gamma^{(1)}$ is a diagonal matrix with diagonal elements being $\beta^{(1)}_1,\cdots,\beta^{(1)}_{{n_1}}$.}} Since $\Gamma^{(1)}$ has full rank, it does not affect the rank of $H_{1}$, and we can treat their product as a {{full rank}} matrix $P^{(1)} = \Gamma^{(1)}H_{1}$. In this way, we can apply the techniques in the linear neural networks to the one with leaky ReLU activation function.


\begin{lemma}\label{Lem:LeakyReLUNNArbitraryLayers}

Let the generative model $G(\cdot):\mathbb{R}^k\to\mathbb{R}^n$ be implemented by a $d$-layer neural network. Let the weight matrix in each layer be defined as in (\ref{Defn:WeightMatrix}). Assume $n_{i+1}\geq n_i, i=1,\cdots,d-1$, $n=n_d$, and
$n_{d-1}\leq n_d-(2l+1)$. Let each entry in each weight matrix $H_{i}$ be drawn i.i.d. randomly according to the standard Gaussian distribution $\mcN(0,1)$. Let the element-wise activation function $\sigma_{i}$ in the $i$-th layer be a leaky ReLU.

Then {{with high probability, the following holds:}} simultaneously for every pair of $\z$ and $\z_0$ such that $\bfz\neq \bfz_0$, we have $\|G(\bfz_0) - G(\bfz)\|_0 \geq 2l+1$, where $G(\cdot)$ is the generative model defined above. $\ell_0$ minimization will be able to recover the ground truth signal if there are no more than $l$ outliers present.  

\end{lemma}


\begin{proof} Let $i$ be the layer index. Take any pair of 
$\bfz'\in \mathbb{R}^{n_{i-1}}$, and $\bfz'_{0}\in \mathbb{R}^{n_{i-1}}$ such that $\bfz'\neq\bfz'_{0}$. By Lemma \ref{Lem:FullRankOfGaussianRandomMatrix}, we know that $H_{i}$ has full rank with probability $1$. Then, with high probability, for every pair of $\bfz'$ and $\bfz'_{0}$ such that $\bfz' \neq \bfz'_{0}$, we have $H_{i}\bfz' \neq H_{i}\bfz'_{0}$. By Lemma \ref{Lem:PatternIrrelevanceInLeakyReLU}, we further have with high probability, for every pair of $\bfz'$ and $\bfz'_{0}$ such that $\bfz' \neq \bfz'_{0}$, $\sigma_{i}(H_{i}(\bfz'-\bfz'_{0}))\neq \bf0$. This means that for a certain layer, with high probability: ``whenever the inputs are different, we will get different outputs". Then by induction, we know that with high probability, for every pair of $\bfz \neq \bfz_{0}$, we will have $\sigma_{d-1}(H_{d-1}\cdots\sigma_{1}(H_{1}\bfz)) \neq \sigma_{d-1}(H_{d-1}\cdots\sigma_{1}(H_{1}\bfz_{0}))$. Therefore, for the last layer, by Lemma \ref{Lem:FullRankAndZeroEntries} and Lemma \ref{Lem:PatternIrrelevanceInLeakyReLU}, the difference between $G(\z_0)$ and $G(\z)$ must be at least $2l+1$ in $\ell_0$ ``norm''.
\end{proof} 






In the following section \ref{Sec:leaky relu} and section \ref{Sec:relu}, we will show the recovery performance guarantees for generative models with leaky ReLU activation function as well as with ReLU activation function. In what follows within these two sections, to simplify our analysis, we consider the case of $M$ being an identity matrix. Namely we observe the whole signal $\boldsymbol{x}$ but some entries of $\boldsymbol{x}$ are corrupted with outliers. Then the optimization problem becomes 
\begin{equation}
\min_{\boldsymbol{z}\in \mathbb{R}^{k}}\|G(\boldsymbol{z})-(G(\bfz_0)+\boldsymbol{e})\|_{1},
\label{optformulation}
\end{equation}
where 
$\bfz_0$ is the ground-truth vector generating the ground-truth signal $G(\bfz_0)$.

In this case, our newly proposed problem can be considered as a generalization of the error correction problem introduced in \cite{candes_decoding_2005}. In \cite{candes_decoding_2005}, the authors considered observing $\bfz_0$ through a single-layer linear generator, but in this paper, we observe $\bfz_0$ through general multi-layer generative models with general non-linear activation functions. 

\section{Recovery Guarantees Using $\ell_{1}$ Minimization for Generative Model with Leaky ReLU activation}\label{Sec:leaky relu}

In this section, we prove the recovery guarantees for signals modeled by generative models with the leaky ReLU activation functions, in the presence of outliers. We need to consider the outputs of the generative model for two different inputs: $G(\boldsymbol{r})=\sigma_{d}(H_{d}\sigma_{d-1}(H_{d-1} \dotsb \sigma_{1}(H_{1}\boldsymbol{r}+\boldsymbol{\delta_{1}})+\dotsb +\boldsymbol{\delta_{d-1}})+\boldsymbol{\delta_{d}})$, and
$G(\boldsymbol{r}+\boldsymbol{c})=\sigma_{d}(H_{d}\sigma_{d-1}(H_{d-1} \dotsb \sigma_{1}(H_{1}(\boldsymbol{r}+\boldsymbol{c})+\boldsymbol{\delta_{1}})+\dotsb +\boldsymbol{\delta_{d-1}})+\boldsymbol{\delta_{d}})$, 
where $\boldsymbol{c} \neq \boldsymbol{0}$ and $\boldsymbol{r}$ are $k$-dimensional vectors, $H_{i}$ is the $i$-th layer's weight matrix  with each element following the standard Gaussian distribution, and $\boldsymbol{\delta_{i}}$ is the $i$-th layer's bias term  each element of which follows the standard Gaussian distribution. We further consider $\sigma_{i}$ as the leaky ReLU activation function in this section.
 Note that when leaky ReLU constant $h=1$, the activation function is linear and the generative model becomes a linear multi-layer neural network. 
 
 Then we present our main theorem of this section as follows, which states that $\ell_1$ minimization offers performance guarantees for generative models. 
 \begin{theorem}(main theorem)\label{leaky_main}
Consider weight matrices $H_{i}^{n_{i}\times n_{i-1}}$ with $i.i.d\ \mathcal{N}(0,1)$ entries, and $\frac{n_{1}}{n_{0}}=\frac{1}{1-\alpha_{1}}$, $\dotsb$, $\frac{n_{d}}{n_{d-1}}=\frac{1}{1-\alpha_{d}}$, where $\alpha_{1}, \dotsb, \alpha_{d}$ are all positive numbers smaller than $1$. The bias vectors $\boldsymbol{\delta_{1}}$, $\boldsymbol{\delta_{2}},\dotsb, \boldsymbol{\delta_{d}}$ have elements drawn i.i.d. from distribution  $\mathcal{N}(0,1)$.
Then there exists a constant $c_{1}>0$ and a constant $\rho^{*}>0$. such that with probability at least $1-e^{-c_{1}n}$, $\boldsymbol{z}=\bfz_0$ is the unique solution to the $\ell_{1}$-minimization problem (\ref{optformulation}) for every ground-truth vector $\bfz_0$ and every error vector $\boldsymbol{e}$ with number of non-zero elements at most $\rho^{*}n$.
\end{theorem}
To prove Theorem \ref{leaky_main}, we need an intermediate Theorem \ref{leaky_conditionforG}, which specifies a condition on the generator $G$ such that the $\ell_1$ minimization can correct sparse outlier errors. 
 \begin{theorem}\label{leaky_conditionforG}
 We have $\boldsymbol{z}=\bfz_0$ is the unique solution to the $\ell_{1}$-minimization problem (\ref{optformulation}) for every ground-truth vector $\bfz_0$ and every error vector $\boldsymbol{e}$ with number of non-zero elements at most $\rho^{*}n$,  if and only if $\|(G(\boldsymbol{r}+\boldsymbol{c})-G(\boldsymbol{r}))_{K}\|_{1}<\|(G(\boldsymbol{r}+\boldsymbol{c})-G(\boldsymbol{r}))_{{K}^\mathsf{c}}\|_{1}$ for every $\boldsymbol{r}\in \mathbb{R}^{k}$ and every non-zero $\boldsymbol{c}\in \mathbb{R}^{k}$, as well as every support $K \subseteq \{1,2,\dotsb,n\}$ with $|K|\leq \rho^{*}n$ ($K^c$ is the complement of $K$).
 \end{theorem}


Now we give the outline of our proof of Theorem \ref{leaky_main}. To prove Theorem \ref{leaky_main},  we will prove the condition in Theorem \ref{leaky_conditionforG} is satisfied for multi-layer neural network based generator $G(\cdot)$.  We first focus on a single layer of the neural network (the last layer), and show that this condition in Theorem \ref{leaky_conditionforG} is satisfied. To achieve this, we use an $\gamma$-net method to cover the unit sphere in the Euclidean space, and show that the condition in  \ref{leaky_conditionforG} is satisfied for every point in the $\gamma$-net. Then we use a continuity technique to extend the result to every possible vector in the Euclidean space. The arguments for the single layer are presented in Lemmas \ref{leaky_lowerbound}, \ref{leaky_intermediate_lowerbound}, and  \ref{leaky_upperbound}. Then we use Lemma \ref{leaky_differentoutputs} to extend the result from a single layer to multiple layers. 

\begin{lemma}\label{leaky_lowerbound}
Consider a matrix $H\in \mathbb{R}^{n \times (n-m)}$ with $i.i.d.\ \mathcal{N}(0,1)$ entries, and let $\alpha=\frac{m}{n}$ be a constant. Then, there exists a constant $\lambda_{min}(\alpha)>0$ and a constant $c_{2}>0$ such that with probability at least $1-e^{-c_{2}n}$, simultaneously for every $\boldsymbol{v} \in \mathcal{S}$, where $\mathcal{S}$ is the unit sphere in $\mathbb{R}^{n-m}$, we have $\|H\boldsymbol{v}\|_{1}>\lambda_{min}(\alpha)n/h$.
\end{lemma}

To prove Lemma \ref{leaky_lowerbound} we need to introduce an intermediate Lemma \ref{leaky_intermediate_lowerbound}.
\begin{lemma}\label{leaky_intermediate_lowerbound}
Given $\alpha$, there exists a constant $\lambda_{max}(\alpha)$ and some constant $c_{3}>0$ such that for every $\boldsymbol{v}\in \mathcal{S}$, $\|H\boldsymbol{v}\|_{1}<\lambda_{max}(\alpha)n/h$ with probability at least $1-e^{-c_{3}n}$.
\end{lemma}
We give proof of Lemma \ref{leaky_intermediate_lowerbound} in appendix.

\begin{proof} (of Lemma \ref{leaky_lowerbound})
For a fixed $\alpha$, we first define
    \begin{equation}
        c_{max}=\frac{1}{n}\max\limits_{\boldsymbol{v}\in \mathcal{S}}\|H\boldsymbol{v}\|_{1}. 
    \end{equation}
And define
\begin{equation}
        c_{min}=\frac{1}{n}\min\limits_{\boldsymbol{v}\in \mathcal{S}}\|H\boldsymbol{v}\|_{1}.
\end{equation}
    Pick a $\gamma$-net $\Sigma_{1}$ of $\mathcal{S}$ with cardinality at most $(1+2/\gamma)^{n-m}$ where $\gamma$ is a small positive number. Then for every $\boldsymbol{v} \in \mathcal{S}$, there exists $\boldsymbol{v'} \in \Sigma_{1}$ such that $\|\boldsymbol{v}-\boldsymbol{v'}\|_{2}\leq \gamma$. Next, define
\begin{equation}
        \theta=\frac{1}{n}\min\limits_{\boldsymbol{v}\in \Sigma_{1}}\|H\boldsymbol{v}\|_{1}.
\end{equation}
     Thus,
\begin{equation}
\begin{aligned}
 \|H\boldsymbol{v}\|_{1}&\geq \|H\boldsymbol{v'}\|_{1}-\|H(\boldsymbol{v}-\boldsymbol{v'})\|_{1}\\
        &\geq \theta n-\gamma c_{max}n
\end{aligned}
\end{equation}
    and we therefore have
\begin{equation}
c_{min}\geq \theta-\gamma c_{max}.
\end{equation}
We want to first show that there exists a positive constant $b$ such that $\theta>b$ with high probability. $i.e. \|H\boldsymbol{v}\|_{1}>bn$ simultaneously $\forall \boldsymbol{v}\in \Sigma_{1}$ with high probability. In fact, by the Chernoff bound, we have
\begin{equation}\label{1}
\begin{aligned}
\mathbb{P}(\theta \leq b)&=\mathbb{P}(\exists \ \boldsymbol{v} \in \Sigma_{1}\  \text{s.t.}\ \|H\boldsymbol{v}\|_{1} \leq bn)\\
&\leq \sum\limits_{\boldsymbol{v}\in \Sigma_{1}}\mathbb{P}(\|H\boldsymbol{v}\|_{1}\leq bn)\\
&\leq (1+2/\gamma)^{n-m}e^{sbn}E[e^{-s\sum\limits_{i}|H_{(i)}\boldsymbol{v}|}]\\
&=e^{((1-\alpha)\text{log}(1+2/\gamma)+\text{log}(E[e^{-s|X|}])+bs)n}, \ \forall s>0
\end{aligned}
\end{equation}
where $X \sim \mathcal{N}(0,1)$ and $H_{(i)}$ is the $i^{th}$ row of $H$. Note that
\begin{equation}
\begin{aligned}
    E[e^{-s|X|}]&=\sqrt{2/\pi}\int^{\infty}_{0}e^{-sx}e^{-\frac{1}{2}x^{2}}dx\\
    &=\frac{1}{s}\sqrt{2/\pi}\int^{\infty}_{0}e^{-y}e^{-\frac{1}{2}(y/s)^{2}}dy\\
    &\leq\frac{1}{s}\sqrt{2/\pi}\int^{\infty}_{0}e^{-y}dy\\
    &=\frac{1}{s}\sqrt{2/\pi}
\end{aligned}
\end{equation}
where we do the substitution $y=sx$ in the second equation.
Also, we will have
\begin{equation}
    E[e^{-s|X|}]\geq \frac{1}{s}\sqrt{2/\pi}\int^{\infty}_{0}e^{-y-\frac{1}{2}y^{2}}dy
\end{equation}
from the second equation, when $s>1$.
By combining the above, we get
\begin{equation}
    E[e^{-s|X|}]=\mathcal{O} (\frac{1}{s}).
\end{equation}
Since (\ref{1}) holds for all $s>0$, we let
\begin{equation}
    s=\gamma^{-(1-\alpha+\epsilon)}
\end{equation} 
for any $\epsilon$ such that $0<\epsilon \leq \alpha$ and let
\begin{equation}
b(\gamma)=\frac{1}{sh}.
\end{equation}
Denote
\begin{equation}
\begin{aligned}
    \kappa(\gamma)&=-(1-\alpha)\text{log}(1+2/\gamma)-\text{log}(E[e^{-s|X|}])-bs\\
    &=-(1-\alpha)\text{log}(1+2/\gamma)-\text{log}(\mathcal{O}(\gamma^{1-\alpha +\epsilon}))-\frac{1}{h}.
\end{aligned}
\end{equation}
Then
\begin{equation}
\mathbb{P}(\theta \leq b(\gamma))\leq e^{-\kappa n}.
\end{equation}
Therefore, for a small enough positive $\gamma$, there exists $\kappa(\gamma)>0$ such that
\begin{equation}
    \mathbb{P}(\theta \leq b(\gamma)=\frac{1}{h}\cdot\gamma^{1-\alpha+\epsilon})\leq e^{-\kappa(\gamma)n}.
\end{equation}
Lemma \ref{leaky_intermediate_lowerbound} implies that there exists a constant $\lambda_{max}(\alpha)$ and some constant $c_{3}$ such that $\mathbb{P}(c_{max}<\lambda_{max}(\alpha))$ is at least $1-e^{-c_{3}n}$.
Now, with $\theta$ and $c_{max}$, we can characterize $c_{min}$. In fact,
\begin{equation}
    \begin{aligned}
    &\mathbb{P}(c_{min}\leq \frac{1}{h}(\gamma^{1-\alpha+\epsilon}-\gamma \lambda_{max}(\alpha)))\\
    \leq &\mathbb{P}(\theta-\gamma c_{max}\leq \frac{1}{h}(\gamma^{1-\alpha +\epsilon}-\gamma \lambda_{max}(\alpha)))\\
    \leq &\mathbb{P}(\theta \leq \frac{1}{h}\cdot\gamma^{1-\alpha +\epsilon})+\mathbb{P}(c_{max}\geq \lambda_{max}(\alpha)/h)\\
    \leq &e^{-\kappa n}+e^{-c_{3}n}
    \end{aligned}
\end{equation}
Then, for a positive $\gamma$ small enough, there exists a constant $c_{2}$ such that
\begin{equation}
    \mathbb{P}(c_{min}\leq \frac{1}{h}(\gamma^{1-\alpha+\epsilon}-\gamma \lambda_{max}(\alpha)))\leq e^{-c_{2}n}.
\end{equation}
Therefore, given $\lambda_{max}(\alpha)$, let
\begin{equation}
    \lambda_{min}(\alpha)=\max\limits_{0<\gamma}(\gamma^{1-\alpha+\epsilon}-\gamma \lambda_{max}(\alpha)),
\end{equation}
where $\gamma$ is small enough. Thus, we have $\lambda_{min}>0$ which ends the proof.
\end{proof}

\begin{lemma}\label{leaky_upperbound}
Consider a matrix $H\in \mathbb{R}^{n \times (n-m)}$ with $i.i.d.\ \mathcal{N}(0,1)$ entries, and let  $\alpha=\frac{m}{n}$ be a constant. Given any $\lambda_{min}(\alpha)>0$, there exists a constant $\rho^{*}(\alpha)>0$ and some constant $c_{4}>0$ such that with probability at least $1-e^{-c_{4}n}$, for every $\boldsymbol{v} \in \mathcal{S}$ in $\mathbb{R}^{n-m}$, and for every set $K \subseteq \{1,2,\dotsb,n\}$, with $|K| \leq \rho^{*}(\alpha)n$, we have $ \|(H\boldsymbol{v})_{K}\|_{1}<\frac{1}{2}\lambda_{min}(\alpha)n$.
\end{lemma}

\begin{proof} (of Lemma \ref{leaky_upperbound})
For any such set $K$ with $|K|=\rho n\ (0<\rho<1)$, we define
\begin{equation}
    d_{max}=\frac{1}{n}\max\limits_{\boldsymbol{v}\in \mathcal{S}}\|(H\boldsymbol{v})_{K}\|_{1}.
\end{equation} 
Given a $\gamma$-net $\Sigma_{3}$ of $\mathcal{S}$ with cardinality at most $(1+2/\gamma)^{n-m}$ and $\gamma$ is a small positive number. We further define
\begin{equation}
    \tau=\frac{1}{n}\max\limits_{\boldsymbol{v}\in \Sigma_{3}}\|(H\boldsymbol{v})_{K}\|_{1}.
\end{equation}
Then for every $\boldsymbol{v}\in \mathcal{S}$, there exists $\boldsymbol{v'}\in \Sigma_{3}$ such that $\|\boldsymbol{v}-\boldsymbol{v'}\|_{2}\leq \gamma$. So for every $\boldsymbol{v}\in \mathcal{S}$, we have
\begin{equation}
    \begin{aligned}
    \|(H\boldsymbol{v})_{K}\|_{1}&\leq \|(H\boldsymbol{v'})_{K}\|_{1}+\|(H(\boldsymbol{v}-\boldsymbol{v'}))_{K}\|_{1}\\
    &\leq \tau n+\gamma d_{max}n
    \end{aligned}
\end{equation}
we therefore have
\begin{equation}
    d_{max}\leq \tau/(1-\gamma).
\end{equation}
Given $\lambda_{min}$, in order to get $\rho^{*}(\alpha)$ such that the lemma holds, we need to find $\rho$ such that for any $K$ with its corresponding $d_{max}$, with overwhelming probability that $d_{max}<\lambda_{min}(1-\gamma)/2$ holds for all $K$ with $|K|=\rho n$ simultaneously. We first look at the probability that $\tau \geq \lambda_{min}(1-\gamma)/2h$ holds for a given set $K$. In fact, by union bound and Chernoff bound, we have the following.
\begin{equation}
\begin{aligned}
&\mathbb{P}(\tau \geq \lambda_{min}(1-\gamma)/2,\  \text{given}\  K)\\
=&\mathbb{P}(\exists \ \boldsymbol{v} \in \Sigma_{3}\  \text{s.t.}\ \|(H\boldsymbol{v})_{K}\|_{1} \geq \lambda_{min}(1-\gamma)n/2)\\
\leq &\sum\limits_{\boldsymbol{v}\in \Sigma_{3}}\mathbb{P}(\|(H\boldsymbol{v})_{K}\|_{1}\geq \lambda_{min}(1-\gamma)n/2)\\
=&\sum\limits_{\boldsymbol{v}\in \Sigma_{3}}\mathbb{P}(\sum\limits_{i\in K}|H_{(i)}\boldsymbol{v}|\geq\lambda_{min}(1-\gamma)n/2)\\
\leq &(1+2/\gamma)^{n-m}\min\limits_{s>0}e^{-s\lambda_{min}(1-\gamma)n/2}E[e^{s\sum\limits_{i\in K}|H_{(i)}\boldsymbol{v}|}]\\
=&(1+2/\gamma)^{(1-\alpha)n}\min\limits_{s>0}e^{-s\lambda_{min}(1-\gamma)n/2}E[e^{s|X|}]^{\rho n}\\
=&e^{((1-\alpha)\text{log}(1+2/\gamma)+\min\limits_{s>0}(\rho \text{log}(E[e^{s|X|}])-s\lambda_{min}(1-\gamma)/2))n},
\end{aligned}
\end{equation}
$\forall t>0$, where $X\sim \mathcal{N}(0,1)$.\\
Given $\rho$, $\lambda_{min}$ and $\gamma$, since the second derivative of $\rho \text{log}(E[e^{s|X|}])-s\lambda_{min}(1-\gamma)/2$ in terms of $s$ is positive, its minimum is achieved where its first derivative is zero. Specifically,
\begin{equation}
    \begin{aligned}
    0&=\frac{d[\rho \text{log}(E[e^{s|X|}])-s\lambda_{min}(1-\gamma)/2]}{ds}\\
    &=\frac{d}{ds}(\rho \text{log}(\sqrt{\frac{2}{\pi}}\int^{\infty}_{0}e^{sx-\frac{1}{2}x^{2}}dx)-s\lambda_{min}(1-\gamma)/2)\\
    &=\frac{\rho\int^{\infty}_{0}xe^{sx-\frac{1}{2}x^{2}}dx}{\int^{\infty}_{0}e^{sx-\frac{1}{2}x^{2}}dx}-\lambda_{min}(1-\gamma)/2
    \end{aligned}
\end{equation}
The solution $s$ is always positive when $\rho<\lambda_{min}(1-\gamma)/(2E[|X|])$. So it is also the solution to $\min\limits_{s>0}(\rho \text{log}(E[e^{s|X|}])-s\lambda_{min}(1-\gamma)/2)$. Next, we consider the probability that $\|(H\boldsymbol{v})_{K}\|_{1}\geq\frac{1}{2}\lambda_{min}n$ for some $\boldsymbol{v}\in \mathcal{S}$ and $K$ with $|K|=\rho n$.
\begin{equation}\label{3}
    \begin{aligned}
    &\mathbb{P}(\exists \boldsymbol{v}\in \mathcal{S}, \exists K \ \text{s.t.}\ |K|=\rho n, \|(H\boldsymbol{v})_{K}\|_{1}\geq \lambda_{min}n/2)\\
    \leq &\binom {n} {\rho n}\mathbb{P}(\exists\boldsymbol{v}\in \mathcal{S}\ \text{s.t.}\ \|(H\boldsymbol{v})_{K}\|_{1}\geq\lambda_{min}n/2,\\
    &(\text{for given}\  K\subseteq \{1,2,...,n\}\ \text{and}\ |K|=\rho n)\\
    =&\binom {n}{\rho n}\mathbb{P}(d_{max}\geq \lambda_{min}/2)\\
    \leq &\binom{n}{\rho n}\mathbb{P}(\tau/(1-\gamma)\geq \lambda_{min}/2)\\
    \leq &2^{nH(\rho)}e^{((1-\alpha)\text{log}(1+2/\gamma)+\min\limits_{s>0}(\rho \text{log}(E[e^{s|X|}])-s\lambda_{min}(1-\gamma)/2))n}\\
    =&e^{(H(\rho)\text{log}2+(1-\alpha)\text{log}(1+2/\gamma)+\min\limits_{s>0}(\rho \text{log}(E[e^{s|X|}])-s\lambda_{min}(1-\gamma)/2))n}
    \end{aligned}
\end{equation}
Since for given $\alpha$ and $\lambda_{min}$, for every $\gamma$, $H(\rho)$ goes to $0$ as $\rho$ goes to $0$, and $\min\limits_{s>0}(\rho \text{log}(E[e^{s|X|}])-s\lambda_{min}(1-\gamma)/2)$ goes to $-\infty$, thus, there exists $\rho(\alpha, \gamma)>0$ such that the exponent of (\ref{3}) is negative for all $\rho\leq\rho(\alpha, \gamma)$. In other words, there exists some constant $c_{4}>0$ such that the probability in (\ref{3}) is no bigger than $e^{-c_{4}n}$ when $\rho=\rho(\alpha, \gamma)$. Then, with probability at least $1-e^{-c_{4}n}$, for every $\boldsymbol{v}\in \mathcal{S}$ and for every set $K\subseteq\{1,2,...,n\}$ with $|K|\leq\rho(\gamma)n$, $\|(H\boldsymbol{v})_{K}\|_{1}<\lambda_{min}n/2$. Then, let $\rho^{*}(\alpha)=\max\limits_{\gamma}\rho(\alpha, \gamma)$, and this ends the proof.
\end{proof}

\begin{lemma}\label{leaky_differentoutputs}
Consider a matrix $H'\in \mathbb{R}^{p\times(p-q)}$ with $i.i.d.\ \mathcal{N}(0,1)$ entries, and let  $\alpha=\frac{q}{p}<1$ be a constant. Let $\boldsymbol{\delta}$ be the bias vector with each element sampled i.i.d. from $\mathcal{N}(0,1)$. Then with high probability, simultaneously for every $ \boldsymbol{u}'\in \mathbb{R}^{p-q}$, and for every $ \boldsymbol{w}'\neq \boldsymbol{0}\in \mathbb{R}^{p-q}$,  $\sigma(H'(\boldsymbol{u}'+\boldsymbol{w}')+\boldsymbol{\delta})\neq \sigma(H'\boldsymbol{u}'+\boldsymbol{\delta})$, where $\sigma$ is the leaky ReLU activation function.
\end{lemma}
\begin{proof} (of Lemma \ref{leaky_differentoutputs})
If $H'$ is a matrix with $i.i.d\  \mathcal{N}(0,1)$ entries, then it has full rank ($p-q$) with high probability. Since $\boldsymbol{w}' \neq \boldsymbol{0}$, with high probability, simultaneously for every $ \boldsymbol{u}'\in \mathbb{R}^{p-q}$, and for every $\boldsymbol{w}'\neq \boldsymbol{0}\in \mathbb{R}^{p-q}$, $H'(\boldsymbol{u}'+\boldsymbol{w}')+\boldsymbol{\delta}\neq H'\boldsymbol{u}'+\boldsymbol{\delta}$. Also, the leaky ReLU activation function $\sigma$ is one-to-one, meaning $\sigma(H'(\boldsymbol{u}'+\boldsymbol{w}')+\boldsymbol{\delta})\neq \sigma(H'\boldsymbol{u}'+\boldsymbol{\delta})$. Thus, with high probability, ``the outputs of a certain layer will always be different, whenever the inputs of this layer are different.''
\end{proof}
Now, equipped with the above four lemmas, we give the proof of the above Theorem \ref{leaky_main}.
\begin{proof} (of Theorem \ref{leaky_main}) We will show that the condition in Theorem \ref{leaky_conditionforG} holds with high probability. For that purpose, we consider two different inputs $\boldsymbol{r}$ and $\boldsymbol{r}+\boldsymbol{c}$, where $\boldsymbol{c}\neq \boldsymbol{0}$. For simplicity, we denote $\boldsymbol{u}=\sigma_{d-1}(H_{d-1} \dotsb \sigma_{1}(H_{1}\boldsymbol{r}+\boldsymbol{\delta_{1}})+ \dotsb +\boldsymbol{\delta_{d-1}})\in \mathbb{R}^{n_{d-1}}$, and denote $\boldsymbol{u}+\boldsymbol{w}=\sigma_{d-1}(H_{d-1} \dotsb \sigma_{1}(H_{1}(\boldsymbol{r}+\boldsymbol{c})+\boldsymbol{\delta_{1}})+ \dotsb +\boldsymbol{\delta_{d-1}})$.

By Lemma \ref{leaky_differentoutputs}, with high probability, the outputs of a certain layer will be different whenever the inputs of that layer are different. Since $\boldsymbol{c}\neq \boldsymbol{0}$, by induction, with high probability ``$\boldsymbol{w}\neq \boldsymbol{0}$ for every $\boldsymbol{r}$ and $\boldsymbol{c} \neq \boldsymbol{0}$.''

Now we lower bound $\|(\sigma_{d}(H_{d}(\boldsymbol{u}+\boldsymbol{w})+\boldsymbol{\delta_{d}})-\sigma_{d}(H_{d}\boldsymbol{u}+\boldsymbol{\delta_{d}})\|_{1}$. In fact, we have
\begin{equation}
    \begin{aligned}
    h\|H_{d}\boldsymbol{w}\|_{1}&=h\|H_{d}(\boldsymbol{u}+\boldsymbol{w})+\boldsymbol{\delta_{d}}-(H_{d}\boldsymbol{u}+\boldsymbol{\delta_{d}})\|_{1}\\
    &\leq \|(\sigma_{d}(H_{d}(\boldsymbol{u}+\boldsymbol{w})+\boldsymbol{\delta_{d}})-\sigma_{d}(H_{d}\boldsymbol{u}+\boldsymbol{\delta_{d}})\|_{1} \\
    &\leq \|H_{d}(\boldsymbol{u}+\boldsymbol{w})+\boldsymbol{\delta_{d}}-(H_{d}\boldsymbol{u}+\boldsymbol{\delta_{d}})\|_{1}\\
    &=\|H_{d}\boldsymbol{w}\|_{1},
    \end{aligned}
\end{equation}
where $h$ is the leaky ReLU constant. By Lemma \ref{leaky_lowerbound}, we know that, with probability at least $1-e^{-c_{2}n}$, for every $\boldsymbol{w} \neq \boldsymbol{0}$,
\begin{equation}
    \|H_{d}\boldsymbol{w}\|_{1}>\|\boldsymbol{w}\|_2\lambda_{min}(\alpha_{d})n/h.
\end{equation}
So with probability at least $1-e^{-c_{2}n}$, for every $\boldsymbol{w} \neq 0$ and every $\boldsymbol{u}$,
\begin{equation}
\begin{split}
    &\|\sigma_{d}(H_{d}(\boldsymbol{u}+\boldsymbol{w})+\boldsymbol{\delta_{d}})-\sigma_{d}(H_{d}\boldsymbol{u}+\boldsymbol{\delta_{d}})\|_{1} \\&\geq h\|H_{d}\boldsymbol{w}\|_{1}>\|\boldsymbol{w}\|_2\lambda_{min}(\alpha_{d})n.
    \end{split}
\end{equation}
By Lemma \ref{leaky_upperbound}, there exists a constant $\rho^*$ such that, with probability at least $1-e^{-c_{4}n}$,  for every subset $K$ with cardinality no bigger than $\rho^* n$, we have
\begin{equation}
     \|(H_{d}\boldsymbol{w})_{K}\|_{1}<\frac{1}{2}\|\boldsymbol{w}\|_2\lambda_{min}(\alpha_{d})n.
\end{equation}
So with probability at least $1-e^{-c_{4}n}$, for every $\boldsymbol{w} \neq 0$ and every $\boldsymbol{u}$,
\begin{equation}
\begin{split}
    &\|(\sigma_{d}(H_{d}(\boldsymbol{u}+\boldsymbol{w})+\boldsymbol{\delta_{d}})-\sigma_{d}(H_{d}\boldsymbol{u}+\boldsymbol{\delta_{d}}))_{K}\|_{1}\\&\leq \|(H_{d}\boldsymbol{w})_{K}\|_{1}<\frac{1}{2}\|\boldsymbol{w}\|_2\lambda_{min}(\alpha_{d})n.
    \end{split}
\end{equation}
In summary, by the union bound, with probability at least $1-e^{-c_{2}n}-e^{-c_{4}n}$,
\begin{equation}
\begin{split}
    &\|(\sigma_{d}(H_{d}(\boldsymbol{u}+\boldsymbol{w})+\boldsymbol{\delta_{d}})-\sigma_{d}(H_{d}\boldsymbol{u}+\boldsymbol{\delta_{d}}))_{K}\|_{1}\\
    <&\|(\sigma_{d}(H_{d}(\boldsymbol{u}+\boldsymbol{w})+\boldsymbol{\delta_{d}})-\sigma_{d}(H_{d}\boldsymbol{u}+\boldsymbol{\delta_{d}}))_{{K}^\mathsf{c}}\|_{1}
    \end{split}
\end{equation}
for every $ \boldsymbol{u} \in \mathbb{R}^{n_{d-1}}$, every $ \boldsymbol{w} \neq \boldsymbol{0}$, and for every set $K$ with $|K|\leq \rho^{*}n$. (Note that there exists a $c_{5}>0$ such that $1-e^{-c_{2}n}-e^{-c_{4}n}\geq 1-e^{-c_{5}n}$.)

Thus there exists a $c_{1}>0$ such that with probability at least $1-e^{-c_{1}n}$, $\|(G(\boldsymbol{r}+\boldsymbol{c})-G(\boldsymbol{r}))_{K}\|_{1}<\|(G(\boldsymbol{r}+\boldsymbol{c})-G(\boldsymbol{r}))_{{K}^\mathsf{c}}\|_{1}$ for every $\boldsymbol{r}\in \mathbb{R}^{k}$ and every non-zero $\boldsymbol{c}\in \mathbb{R}^{k}$, and every subset $K$ with $|K|\leq \rho^{*}n$. Then by Theorem \ref{leaky_conditionforG}, with probability at least $1-e^{-c_{1}n}$,  the $\ell_1$ minimization can correctly recover the ground-truth signal $\bfz_0$  (no matter what value this $\bfz_0$ takes) under every error vector with support size at most $\rho^{*}n$.
\end{proof}


\section{Recovery Guarantees Using $\ell_{1}$ Minimization for Generative Model with ReLU activation}\label{Sec:relu}

In this section, we derive the recovery guarantee for generative models with ReLU activation functions, with observations corrupted by outliers. 
The technical challenge in performing analysis for the ReLU activation function is that two different inputs to a ReLU activation function can lead to the same output, making it challenging to lower bound the difference between the outputs of the multiple-layer generative model for two different inputs. However, we are able to provide performance guarantees via tailored analysis  utilizing the piecewise linear property of the ReLU activation function. The main results for the ReLU activation function are given in Theorem \ref{relu_main}. 

\begin{theorem}(main theorem)\label{relu_main}
Let $\bfz_0$ be a vector from $\mathbb{R}^{n_0}$, namely  from $\mathbb{R}^{k}$. Consider weight matrices $H_{i}^{n_{i}\times n_{i-1}}$ with $i.i.d\ \mathcal{N}(0,1)$ entries, denote that $\frac{n_{1}}{n_{0}}=\frac{1}{1-\alpha_{1}}$, $\dotsb$, $\frac{n_{d}}{n_{d-1}}=\frac{1}{1-\alpha_{d}}$, where $\alpha_{1}, \dotsb, \alpha_{d}$ are all positive constant numbers strictly smaller than but close enough to $1$. The bias vectors $\boldsymbol{\delta_{1}}, \boldsymbol{\delta_{2}}, \dotsb, \boldsymbol{\delta_{d}}$ have elements drawn $i.i.d.$ from distribution $\mathcal{N}(0,1)$. Then there exists a constant $c_{6}>0$ and a constant $\rho^{*}>0$ such that with probability at least $1-e^{-c_{6}n}$, $\boldsymbol{z}=\bfz_0$ is the unique solution to the $\ell_{1}$-minimization problem (\ref{optformulation}) for every error vector $\boldsymbol{e}$ with the number of non-zero elements at most $\rho^{*}n$.
\end{theorem}

 We remark that in Theorem \ref{relu_main}, the constants $\alpha_i$'s are required to be close to 1. Namely, for every $i$, we require $1-\alpha_i<\epsilon$, for a constant $\epsilon>0$.  Similar to the last section, in order to prove our main theorem (Theorem \ref{relu_main}) of this section, we introduce the following intermediate Theorem \ref{relu_conditionforG} which gives the condition on the generator $G$ such that the $\ell_{1}$minimization can correct sparse outlier errors. This condition works for any given value of $\bfz_0$,  instead of \emph{simultaneously} working for all the possible values of the input $\bfz_0$.

\begin{theorem}\label{relu_conditionforG}
 We have $\boldsymbol{z}=\bfz_0$ is the unique solution to $\ell_{1}$ minimization problem (\ref{optformulation}) for every error vector $\boldsymbol{e}$ with number of non-zero elements at most $\rho^{*} n$ if and only if  $\|(G(\bfz_0+\boldsymbol{c})-G(\bfz_0))_{K}\|_{1}< \|(G(\bfz_0+\boldsymbol{c})-G(\bfz_0))_{{K}^\mathsf{c}}\|_{1}$,  for every non-zero $\boldsymbol{c}\in \mathbb{R}^{k}$ as well as every support $K \subseteq \{1,2,\dotsb, n\}$ with $|K|\leq \rho^{*} n$.
 \end{theorem}

 So we just need to prove the condition in Theorem \ref{relu_conditionforG} holds for the considered multi-layer generative models with the regular ReLU activation functions. Towards this end, we first focus on a single layer of the neural network (the last layer) and show that the condition in Theorem \ref{relu_conditionforG} is satisfied (as stated and proved in Theorem \ref{keylemma_newbyweiyu}). To prove Theorem \ref{keylemma_newbyweiyu}, we use the sphere covering method in the Euclidean space. Compared with the leaky ReLU,  for the regular ReLU we do not have a nice lower bound on the $\ell_1$ norm of the change in the outputs when we consider two different inputs to the generator neural network. We overcome this difficulty non-trivially by proving the concentration of the right derivative of the $\ell_1$ norm of the change with respect to the magnitude of the difference between two inputs (as shown in 
 Lemmas \ref{relu_lowerbound}, \ref{relu_lowerbound_union bound}, \ref{lowerbound_uniform_overeveryc}).

 
 We now introduce a theorem concerning one layer of the generative model for two different inputs. 
\begin{theorem}\label{keylemma_newbyweiyu}
Consider a matrix $H \in \mathbb{R}^{n\times(n-m)}$ with $i.i.d \ \mathcal{N} (0,1)$ entries, and let $\alpha=\frac{m}{n}<1$ be a constant close enough to $1$. Let the bias vector $\boldsymbol{\delta}$ have elements drawn $i.i.d.$ from distribution $\mathcal{N}(0,1)$. We consider any vector $\boldsymbol{u} \in \mathbb{R}^{n-m}$. Then there exists a constant $\lambda_{min}(\alpha)>0$ and a positive constant $\rho^*(\alpha)>0$, such that, with high probability, simultaneously for every $\boldsymbol{w} \in \mathbb{R}^{n-m} \neq \boldsymbol{0}$, and for every set $K\subseteq \{1,2, \dotsb, n\}$ with $|K|\leq \rho^{*}(\alpha)n$, we have $\|\sigma(H\boldsymbol{u}+\boldsymbol{\delta})-\sigma(H(\boldsymbol{u}+\boldsymbol{w})+\boldsymbol{\delta})\|_{1}> \lambda_{min} (\alpha)n\|\boldsymbol{w}\|_2 $ and 
$\|(\sigma(H\boldsymbol{u}+\boldsymbol{\delta})-\sigma(H(\boldsymbol{u}+\boldsymbol{w})+\boldsymbol{\delta}) )_{K}\|_{1}<\frac{1}{2}\lambda_{min}(\alpha)n\|\boldsymbol{w}\|_2$.
\end{theorem}

\begin{proof} (of Theorem \ref{keylemma_newbyweiyu})  We can write
\[
H(\boldsymbol{u}+\boldsymbol{w})+\boldsymbol{\delta} = 
\left[
  \begin{array}{cc}
    H & \boldsymbol{\delta} \\
  \end{array}
\right]
\begin{bmatrix}
           \boldsymbol{u}+\boldsymbol{w} \\
           1 \\
         \end{bmatrix},
\]
\[
H\boldsymbol{u}+\boldsymbol{\delta} = 
\left[
  \begin{array}{cc}
    H & \boldsymbol{\delta} \\
  \end{array}
\right]
\begin{bmatrix}
           \boldsymbol{u} \\
           1 \\
         \end{bmatrix}.
\]
We let $\boldsymbol{u}'=\sqrt{\|\boldsymbol{u}\|^2+1}   [1\quad0 \quad \dotsb \quad 0]^{T}$ and $\boldsymbol{u}'+\boldsymbol{w}'=\sqrt{\|\boldsymbol{u}\|^2+1} [1\quad0 \quad \dotsb \quad 0]^{T}+\sqrt{\|\boldsymbol{u}\|^2+1} t\boldsymbol{c}'$,
where  $\boldsymbol{u}'$, $\boldsymbol{w}'$ and $\boldsymbol{c}'$  are vectors of dimension $(n-m+1) \times 1$,  $\|\sqrt{\|\boldsymbol{u}\|^2+1} t\boldsymbol{c}'\|_2= \|\boldsymbol{w}\|_2$,      $\boldsymbol{c}'\in \mathbb{R}^{n-m+1}$ is on the unit sphere $\mathcal{S}$ and $t$ is a positive number. From the rotational invariance of the Gaussian distributions, we have
\[
H(\boldsymbol{u}+\boldsymbol{w})+\boldsymbol{\delta} = 
H'
(\boldsymbol{u}'+\boldsymbol{w}'), 
\]
\[
H\boldsymbol{u}+\boldsymbol{\delta} =H'
\boldsymbol{u}', 
\]
where $H'$ is an  $n \times (n-m+1)$ random matrix with elements i.i.d. sampled from the standard Gaussian distribution $\mathcal{N} (0,1)$.    

Let $\boldsymbol{h}$ be the first column of $H'$.  Because of the positive proportional property of the ReLU function (the output of a ReLU function is also multiplied by the same $a$ if the input is multiplied by a positive number $a>0$),   proving Theorem \ref{keylemma_newbyweiyu} can be achieved by proving the following Theorem \ref{simultaneously}.
\end{proof}
\begin{theorem}\label{simultaneously}
Consider a matrix $H' \in \mathbb{R}^{n\times(n-m+1)}$ with $i.i.d \ \mathcal{N} (0,1)$ entries, let $\alpha=\frac{m}{n}<1$ but close enough to $1$, there exists a constant $\lambda_{min}(\alpha)>0$ and a constant $\rho^*>0$, such that, with high probability, simultaneously for every $\boldsymbol{c}'\in \mathcal{S} \subset \mathbb{R}^{n-m}$,  for every $t>0$, and for every set $K\subseteq \{1,2, \dotsb, n\}$, with $|K|\leq \rho^{*}n$, we have $\|\sigma(\boldsymbol{h}+tH'\boldsymbol{c}')-\sigma(\boldsymbol{h})\|_{1}>\lambda_{min}(\alpha)nt$ and $\|(\sigma(\boldsymbol{h}+tH'\boldsymbol{c}')-\sigma(\boldsymbol{h}))_{K}\|_{1}<\frac{1}{2}\lambda_{min}(\alpha)nt$.
\end{theorem}
\begin{proof}(of Theorem \ref{simultaneously})
In order to prove Theorem \ref{simultaneously}, we will need the result from the following Lemma \ref{slope_change}.
\begin{lemma}\label{slope_change}
For a fixed $\boldsymbol{c}'$, the right derivative (slope) of $\|\sigma(\boldsymbol{h}+tH'\boldsymbol{c}')-\sigma(\boldsymbol{h})\|_{1}$ with respect to $t$ only changes at a finite number of points.
\end{lemma}
\begin{proof} (of Lemma \ref{slope_change}) The activation function $\sigma$ is a piece-wise linear function, so the function $\|\sigma(\boldsymbol{h}+tH\boldsymbol{c}')-\sigma(\boldsymbol{h})\|_{1}$ is also a piece-wise linear function in terms of $t$, and thus only changes its right derivative at finitely many points. 
\end{proof}

We further introduce Lemmas \ref{relu_lowerbound} and \ref{relu_upperbound} to lower bound $\|\sigma(\boldsymbol{h}+tH\boldsymbol{c}'+\boldsymbol{\delta})-\sigma(\boldsymbol{h}+\boldsymbol{\delta})\|_{1}$ and upper bound $\|(\sigma(\boldsymbol{h}+tH\boldsymbol{c}'+\boldsymbol{\delta})-\sigma(\boldsymbol{h}+\boldsymbol{\delta}))_{K}\|_{1}$.

\begin{lemma} \label{relu_lowerbound}
Consider a matrix $H'\in \mathbb{R}^{n \times(n-m+1)}$ with $i.i.d\  \mathcal{N}(0,1)$ entries, let $\alpha=\frac{m}{n}<1$ be a positive constant close enough to $1$, and let $\boldsymbol{h}$ be the first column of $H'$. 
Then there exists a constant $\lambda_{min}(\alpha)>0$ and a constant $c_{7}>0$ such that, for any fixed $t> 0$ and any fixed $\boldsymbol{c}' \in \mathcal{S}$ ($\mathcal{S}$ is the unit sphere in $\mathbb{R}^{n-m+1}$), with probability at least $1-e^{-c_{7}n}$, the right derivative of $\|\sigma(\boldsymbol{h}+tH'\boldsymbol{c}')-\sigma(\boldsymbol{h})\|_{1}$ with respect to $t$ is at least $\lambda_{min}(\alpha) n$.
\end{lemma}

Please refer to Appendix for proof of Lemma \ref{relu_lowerbound}.




By Lemma \ref{slope_change}, the right derivative of the function $\|\sigma(\boldsymbol{h}+tH'\boldsymbol{c}')-\sigma(\boldsymbol{h})\|_{1}$ only changes at finite number of points. Without loss of generality, we consider the change point $t_{1}$ at which the right derivative of $|\sigma(\boldsymbol{h}_{(1)}+tH'_{(1)}\boldsymbol{c}')-\sigma(\boldsymbol{h}_{(1)})|$ changes, where $\boldsymbol{h}_{(i)}$ is the $i$-th element of $\boldsymbol{h}$ and $H'_{(i)}$ is the $i$-th row of $H'$.  Notice that $t_1$ is independent of the other rows of $H'$. For this fixed $t_{1}$ (from the perspective of the other rows of $H'$), by Lemma \ref{relu_lowerbound}, with high probability, the right derivative of $\|\sigma(\boldsymbol{h}+tH'\boldsymbol{c}')-\sigma(\boldsymbol{h})\|_{1}$ at $t_1$, denoted by $D(\|\sigma(\boldsymbol{h}+tH'\boldsymbol{c}')-\sigma(\boldsymbol{h})\|_{1})$, satisfies 
\begin{equation}
\begin{split}
&D (\|\sigma(\boldsymbol{h}+tH'\boldsymbol{c}')-\sigma(\boldsymbol{h})\|_{1})\\
\geq &D(\|\sum\limits_{i\neq1}\sigma(\boldsymbol{h}_{(i)}+tH'_{(i)}\boldsymbol{c}')-\sigma(\boldsymbol{h}_{(i)})\|_{1})\\
>&(n-1) \lambda_{min} (\frac{m-1}{n-1}).
\end{split}
\end{equation}
The same holds true at other slope changing points $t_{i}$, where $i=1, 2, \dotsm, n_{d}$. By the union bound over the $n$ changing points, we have the following Lemma \ref{relu_lowerbound_union bound}. 
\begin{lemma} \label{relu_lowerbound_union bound}
Consider a matrix $H'\in \mathbb{R}^{n \times(n-m+1)}$ with $i.i.d\  \mathcal{N}(0,1)$ entries, let $\alpha=\frac{m}{n}<1$ but close to $1$ be a constant, and let $\boldsymbol{h}$ be the first column of $H'$.  Let $\boldsymbol{c}' \in \mathcal{S}$ ($\mathcal{S}$ is the unit sphere in $\mathbb{R}^{n-m+1}$) be a fixed vector. Then there exists a constant $\lambda_{min}(\alpha)>0$ \footnote{$\lambda_{min}(\alpha)$'s may be different across different lemmas in this section. But they are all positive constants, so we reuse this notation to reduce the number of introduced quantities.} and some constant $c_{8}$ such that, with probability $1-e^{-c_{8}n}$, simultaneously for every $t\geq 0$, the right 
 derivative of $\|\sigma(\boldsymbol{h}+tH'\boldsymbol{c}')-\sigma(\boldsymbol{h})\|_{1}$ is at least $\lambda_{min}(\alpha) n$, and, moreover, $\|\sigma(\boldsymbol{h}+tH'\boldsymbol{c}')-\sigma(\boldsymbol{h})\|_{1}\geq\lambda_{min}(\alpha)nt$.
\end{lemma}

We note that for any $\boldsymbol{c}' \in \mathcal{S}$ and $\boldsymbol{c}'' \in \mathcal{S}$,
\begin{equation}
\begin{split}
&\left|\|\sigma(\boldsymbol{h}+tH'\boldsymbol{c}')-\sigma(\boldsymbol{h})\|_{1} -\|\sigma(\boldsymbol{h}+tH'\boldsymbol{c}'')-\sigma(\boldsymbol{h})\|_{1}\right|\\
\leq &t\|H'(\boldsymbol{c}'-\boldsymbol{c}'')\|_1\\
\leq &\lambda_{max}(\alpha)\|\boldsymbol{c}'-\boldsymbol{c}''\|_2 nt,
\end{split}
\end{equation}
according to Lemma \ref{leaky_intermediate_lowerbound} (taking $h$ there equal to $1$). 
So, through a sphere covering argument (taking the union bound over the $(1+\frac{2}{\gamma})^{(n-m)}=(1+\frac{2}{\gamma})^{(1-\alpha) n}$ discrete points covering the unit sphere as in the proof of Lemmas \ref{leaky_intermediate_lowerbound} and \ref{leaky_upperbound}) over the unit sphere $\mathcal{S}$, we obtain the following Lemma \ref{lowerbound_uniform_overeveryc} lower bounding $\|\sigma(\boldsymbol{h}+tH'\boldsymbol{c}')-\sigma(\boldsymbol{h})\|_{1}$ \emph{uniformly and simultaneously} for every $\boldsymbol{c}'$ belonging to $\mathcal{S}$ and every $t\geq 0$.
\begin{lemma} \label{lowerbound_uniform_overeveryc}
Consider a matrix $H'\in \mathbb{R}^{n \times(n-m+1)}$ with $i.i.d\  \mathcal{N}(0,1)$ entries. Let $\alpha=\frac{m}{n}<1$ be a constant close enough to 1, and let $\boldsymbol{h}$ be the first column of $H'$.  Then there exists a constant $\lambda_{min}(\alpha)>0$ and some constant $c_{9}$ such that, with probability $1-e^{-c_{9}n}$, the following statement holds true: ``Simultaneously for every $t\geq 0$, and simultaneously for every $\boldsymbol{c}' \in \mathcal{S}$ ($\mathcal{S}$ is the unit sphere in $\mathbb{R}^{n-m+1}$),  $\|\sigma(\boldsymbol{h}+tH'\boldsymbol{c}')-\sigma(\boldsymbol{h})\|_{1}\geq\lambda_{min}(\alpha)nt$.''
\end{lemma}

Notice that $\|(\sigma(\boldsymbol{h}+tH'\boldsymbol{c}')-\sigma(\boldsymbol{h}))_K\|_{1} \leq t\|(H'\boldsymbol{c}')_K\|_{1}$, where $K$ is a subset of $\{1, 2, \dotsb, n\}$. Using Lemma \ref{leaky_upperbound}, we thus have an upper bound for $\|\sigma(\boldsymbol{h}+tH'\boldsymbol{c}')-\sigma(\boldsymbol{h})\|_{1}$ over a subset as presented in Lemma \ref{relu_upperbound}. 

\begin{lemma}\label{relu_upperbound}
Consider a matrix $H'\in \mathbb{R}^{n \times (n-m+1)}$ with $i.i.d\  \mathcal{N}(0,1)$ entries. Let $\alpha=\frac{m}{n}$ be a constant, $\boldsymbol{h}$ be the first column of $H'$. Let $\lambda_{min}(\alpha)>0$ be a constant. Then there exists a constant $\rho^{*}(\alpha)>0$ and some constant $c_{10}>0$ such that with probability at least $1-e^{-c_{10}n}$, simultaneously for every $\boldsymbol{c}' \in \mathcal{S}$ in $\mathbb{R}^{n-m+1}$, for every $t\geq 0$, and for every set $K\subseteq \{1,2,\dotsb, n\}$ with $|K| \leq \rho^{*}(\alpha)n$, we have $ \|(\sigma(\boldsymbol{h}+tH'\boldsymbol{c}')-\sigma(\boldsymbol{h}))_{K}\|_{1}<\frac{1}{2}\lambda_{min}(\alpha)nt$, where $\sigma$ is the ReLU activation function.
\end{lemma}

\begin{proof} (for Lemma \ref{relu_upperbound})
Since $\sigma$ is the ReLU activation function, $\|(\sigma(\boldsymbol{h}+tH'\boldsymbol{c}')-\sigma(\boldsymbol{h}))_K\|_{1} \leq \|(\boldsymbol{h}+tH'\boldsymbol{c}'-\boldsymbol{h})_K\|_{1}= t\|(H'\boldsymbol{c}')_K\|_{1}$. By Lemma \ref{leaky_upperbound}, the conclusion holds.
\end{proof}


Combining Lemma \ref{lowerbound_uniform_overeveryc} and Lemma \ref{relu_upperbound}, we complete the proof of Theorem \ref{simultaneously}.
\end{proof}

As discussed in the proof of Theorem \ref{keylemma_newbyweiyu}, Theorem \ref{simultaneously} implies Theorem \ref{keylemma_newbyweiyu}.

By Theorems \ref{relu_conditionforG} and \ref{keylemma_newbyweiyu}, we can prove Theorem \ref{relu_main}.

\begin{proof} (of Theorem \ref{relu_main}) We will show that the condition in Theorem \ref{relu_conditionforG} holds with high probability. We consider two different inputs $\bfz_0$ and $\bfz_0+\boldsymbol{c}$, where $\boldsymbol{c} \neq \boldsymbol{0}$. We denote $\boldsymbol{u}'=\sigma_{d-1}(H_{d-1} \dotsb \sigma_{1}(H_{1}\bfz_0+\boldsymbol{\delta_{1}})+ \dotsb +\boldsymbol{\delta_{d-1}})\in \mathbb{R}^{n_{d-1}}$ and denote $\boldsymbol{u}'+\boldsymbol{w}'=\sigma_{d-1}(H_{d-1} \dotsb \sigma_{1}(H_{1}(\bfz_0+\boldsymbol{c})+\boldsymbol{\delta_{1}})+ \dotsb +\boldsymbol{\delta_{d-1}})\in \mathbb{R}^{n_{d-1}}$. Theorem \ref{keylemma_newbyweiyu} implies that for this fixed $\bfz_0$, with high probability, simultaneously for every $\boldsymbol{c}\neq \boldsymbol{0}$, we have $\sigma_{1}(H_{1}(\bfz_0+\boldsymbol{c})+\boldsymbol{\delta_{1}}) \neq \sigma_{1}(H_{1}\bfz_0+\boldsymbol{\delta_{1}})$. By induction, with high probability,  $\boldsymbol{w}' \neq \boldsymbol{0}$. It also follows from Theorem \ref{keylemma_newbyweiyu} that, with high probability, we have $\|(G(\bfz_0+\boldsymbol{c})-G(\bfz_0))_{K}\|_{1}<\| (G(\bfz_0+\boldsymbol{c})-G(\bfz_0))_{{K}^\mathsf{c}}\|_{1}$ simultaneously for every $\boldsymbol{c}\neq \boldsymbol{0}$. Then by Theorem \ref{relu_conditionforG}, there is a $c_{6}>0$ such that with probability at least $1-e^{-c_{6}n}$, the formulation (\ref{optformulation}) succeeds in recovering signals with an error vector whose number of non-zero elements is at most $\rho^{*}n$.



\end{proof}

\section{Numerical Results}\label{Sec:NumericalResults}
In this section, we present experimental results to validate our theoretical analysis. We first { introduce} the datasets and generative model used in our experiments. Then, we { present simulation} results { verifying} our proposed { outlier detection algorithms.}
  
{We use two datasets MNIST\cite{Lecun:1998} and CelebFaces Attribute (CelebA) \cite{Liu:2015} in our experiments. For generative models, we use variational auto-encoder (VAE) \cite{kingma_auto-encoding_2014} and deep convolutional generative adversarial networks (DCGAN) \cite{Goodfellow:2014,Goodfellow:2016,Radford:2015}.}

\subsection{MNIST and VAE}

The MNIST database contains approximately 70,000 images of handwritten digits which are divided into two separate sets: training and test sets. The training set has 60,000 examples and {the} test set has 10,000 examples. Each image {is of size} 28 $\times$ 28 where each pixel has values within $[0,1]$. We { adopt the pre-trained VAE generative model in \cite{bora_compressed_2017} for MNIST dataset.
The VAE system consists of an encoding network and a decoding network in the inverse architecture.} Specifically, the encoding network has fully connected structure with input, output and two hidden layers. The input has the vector size of 784 elements, the output produces the latent vector having the size of 20 elements. Each hidden layer possesses 500 neurons. The fully connected network creates 784 -- 500 -- 500 -- 20 system. Inversely, the decoding network consists of three fully connected layers converting from latent representation space to image space, i.e., 20--500--500--784 network. { In the training process, the decoding network is trained in such a way to produce the similar distribution of original signals. Training dataset of 60,000 MNIST image samples was used to train the VAE networks. The training process was conducted by Adam optimizer \cite{kingma_adam:_2014} with batch size 100 and learning rate of 0.001.} When the VAE is well-trained, we exploit the decoding neural network as a generator to produce input-like signal $\hat{\x} = G(\z)\in\mathbb{R}^n$. This processing maps a low dimensional signal $\z\in\mathbb{R}^k$ to a high dimensional signal $G(\z)\in\mathbb{R}^n$ such that $\|G(\z) - \x\|_2$ can be small, where the $k$ is set to be $20$ and $n=784$.

Now, we solve the outlier detection problem to find the optimal latent vector $\bf z$ and its corresponding representation vector $G(\z)$ by $\ell_1$-norm and { $\ell_2$-norm minimizations} which are explained as follows. For $\ell_1$ minimization, we consider both the case where we solve (\ref{Defn:L1Minimization}) by the ADMM algorithm introduced in Section \ref{Sec:ADMMAlgorithm}, and the case where we solve (\ref{Defn:SquaredL1Minimization}) by gradient descent (GD) algorithm \cite{Barzilai88}. {  We note that the optimization problem in (\ref{Defn:L1Minimization}) is non-differentiable with respect to $\bf z$, thus, it cannot be solved by GD algorithm. The optimization problem in (\ref{Defn:SquaredL1Minimization}), however, can be solved in practice using GD algorithm. It is because (\ref{Defn:SquaredL1Minimization}) is only non-differentiable at a finite number of points. Therefore, (\ref{Defn:SquaredL1Minimization}) can be effectively solved by a numerical GD solver. }{ In what follows,} we will simply refer to the former case (\ref{Defn:L1Minimization}) as { $\ell_1$ ADMM} and the latter case (\ref{Defn:SquaredL1Minimization}) as { $(\ell_1)^2$ GD}. For $\ell_2$ minimization, we solve the optimization problems (\ref{Defn:SquaredL2Minimization}), and (\ref{Defn:RegularizedSquaredL2Minimization}). 
Both (\ref{Defn:SquaredL2Minimization}) and (\ref{Defn:RegularizedSquaredL2Minimization}) are solved by GD algorithm in \cite{Barzilai88}. {We} will refer to the former case (\ref{Defn:SquaredL2Minimization}) as { $(\ell_2)^2$ GD}, and the later (\ref{Defn:RegularizedSquaredL2Minimization}) as { $(\ell_2)^2$ GD + reg}. The regularization parameter $\lambda$ in (\ref{Defn:SquaredL2Minimization}) is set to be $0.1$.

\subsection{CelebA and DCGAN}

The CelebFaces Attribute dataset consists of over 200,000 celebrity images. We first resize the image samples to $64\times 64$ RGB images, and each image will have {in total} $12288$ pixels with values from $[-1,1]$. { We adopt the pre-trained deep convolutional generative adversarial networks (DCGAN) in \cite{bora_compressed_2017} to conduct the experiments on CelebA dataset}. Specifically, DCGAN consists of a generator and a discriminator. Both generator and discriminator have the structure of one input layer, one output layer and 4 convolutional hidden layers. We map the vectors from latent space of size $k$ = 100 elements to signal space vectors having the size of $n=64\times64\times3= 12288$ elements.

In the training process, { both the generator and the discriminator were trained}. While training the discriminator to identify the {fake} images generated from the generator and ground truth images, the generator is also trained to increase the quality of its fake images so that the discriminator cannot identify the generated samples and ground truth image. This idea {is} based on Nash equilibrium \cite{Nash48} of game theory. { The DCGAN was trained with 200,000 images from CelebA dataset.  We further use additional 64 images for testing our algorithms.} The training process was conducted by the Adam optimizer \cite{kingma_adam:_2014} with learning rate 0.0002,
momentum $\beta_1 = 0.5$ and batch size of 64 images.

Similarly, we solve the outlier detection problems in (\ref{Defn:L1Minimization}), (\ref{Defn:SquaredL1Minimization}), (\ref{Defn:SquaredL2Minimization}) and (\ref{Defn:RegularizedSquaredL2Minimization}) by { $\ell_1$ ADMM, $(\ell_1)^2$ GD}, { $(\ell_2)^2$ GD} and { $(\ell_2)^2$ GD + reg} algorithms, respectively. The regularization parameter $\lambda$ in (\ref{Defn:RegularizedSquaredL2Minimization}) is set to be $0.001$. Besides, we also solve the outlier detection problem by Lasso on the images in both the discrete cosine transformation domain (DCT) \cite{Ahmed:74} and the wavelet transform domain (Wavelet) \cite{Daubechies:88}.

\subsection{Experiments and Results}

\subsubsection{Reconstruction with various numbers of measurements}

The theoretical result in \cite{bora_compressed_2017} showed that a random Gaussian measurement $ \mcM$ is effective for signal reconstruction in CS. Therefore, for evaluation, we set $ \mcM$ as a random matrix with i.i.d. Gaussian entries with zero mean and { standard deviation of 1}. Also, every element of noise vector $\eta$ is an i.i.d. Gaussian random variable.
In our experiments, we carry out the performance comparisons between our proposed $\ell_1$ minimization (\ref{Defn:L1Minimization}) with ADMM algorithm (referred to as { $\ell_1$ ADMM}) approaches, { $(\ell_1)^2$} minimization (\ref{Defn:SquaredL1Minimization}) with GD algorithm (referred to as { $(\ell_1)^2$ GD}), regularized { $(\ell_2)^2$ minimization} (\ref{Defn:RegularizedSquaredL2Minimization}) with GD algorithm (referred to as{ $(\ell_2)^2$ GD + reg}), $(\ell_2)^2$ minimization (\ref{Defn:SquaredL2Minimization}) with GD algorithm (referred as { $(\ell_2)^2$ GD}), Lasso in DCT domain, and Lasso in wavelet domain \cite{bora_compressed_2017}. We use the reconstruction error as our performance metric, which is defined as: $\rm{error} = \| \hat{\bf x} - {\bf x}_0 \|_2^2$, where $\hat{\bf x}$ is an estimate of ${\bf x}_0$ returned by the algorithm. 

For MNIST, we set the standard deviation of the noise vector so that $\sqrt{\mathbb{E}[\|\bf \eta \|^2]} = 0.1$. We conduct $10$ random restarts with $1000$ steps per restart and pick the reconstruction with best measurement error. For CelebA, we set the standard deviation of entries in the noise vector so that $\sqrt{\mathbb{E}[\|\bf \eta \|^2]} = 0.01$. We conduct $2$ random restarts with $500$ update steps per restart and pick the reconstruction with best measurement error.

Fig.2a and Fig.2b show the performance of CS system in the presence of 3 outliers. We compare the reconstruction error versus {the} number of measurements for the $\ell_1$-minimization based methods with ADMM (\ref{Defn:L1Minimization}) and GD (\ref{Defn:SquaredL1Minimization}) algorithms in Fig.2a. In Fig.2b, we plot the reconstruction error versus the number of measurements for $\ell_2$-minimization based methods with and without regularization in (\ref{Defn:SquaredL2Minimization}) and (\ref{Defn:RegularizedSquaredL2Minimization}), respectively. {The outliers' values and positions are randomly {{generated}}. To generate the outlier vector, we first create a vector $\bfe \in {\R^m}$ having all zero elements. Then, we randomly generate $l$ integers in the range $[1,m]$ (following uniform distribution) indicating the positions of outlier in vector $\bfe$. Now, for each generated outlier position, we assign a large random value uniformly distributed in the range [5000, 10000]. The outlier vector $\bfe$ is then added into CS model.} As we can see in Fig.2a when $\ell_1$ minimization algorithms { ($\ell_1$ ADMM and $(\ell_1)^2$ GD)} were used, the reconstruction errors fast converge to low values as the number of measurements increases. On the other hand, the $\ell_2$-minimization based recovery do not {{work well}} as can be seen in  Fig.2b even when number of measurements increases to a large quantity. One observation can be made from Fig.2a is that after $100$ measurements, our algorithm's performance saturates, higher number of measurements does not enhance the reconstruction error performance. This is because of the limitation of VAE architecture.

Similarly, we conduct our experiments with CelebA dataset  using {the} DCGAN generative model. In Fig. 3a, we show the reconstruction error {{change}} as we increase the number of measurements both for  $\ell_1$-minimization based ``{$\ell_1$ ADMM'' and ``$(\ell_1)^2$ GD'' algorithms}. In Fig.3b, we compare reconstruction errors of $\ell_2$-minimization using GD algorithms and Lasso { with the DCT and wavelet bases}. We observed that {in} the presence of outliers, $\ell_1$-minimization based methods outperform the $\ell_2$-minimization based  algorithms and methods using Lasso.

We plot the sample reconstructions by Lasso and our algorithms in Fig. 4. We observed that { $\ell_1$ ADMM and $(\ell_1)^2$ GD} approaches are able to reconstruct the images with only as few as 100 measurements while conventional Lasso and $\ell_2$ minimization algorithm are unable to do the same when number of outliers is as small as 3.

For CelebA dataset, we first show the {{reconstruction performance}} when the number of outlier is 3 and {{the}} number of measurements is 500. We plot the reconstruction results using {$\ell_1$ ADMM and $(\ell_1)^2$ GD} algorithms with DCGAN in Fig. 5. Lasso DCT and Wavelet, { $(\ell_2)^2$ GD and $(\ell_2)^2$ GD + reg} with DCGAN reconstruction results are showed in Fig. 6. Similar to the result in MNIST set, the proposed  $\ell_1$-minimization based algorithms with DCGAN perform better in the presence of outliers. This is because $\ell_1$-minimization-based approaches can successfully eliminate the outliers while $\ell_2$-minimization-based methods do not.

In Figures 5 and 6, we provide further numerical results for outlier detection using $\ell_1$ based algorithms and $\ell_2$ based algorithms for CelebA datasets. We can see that $\ell_1$ minimization based methods perform much better than the $\ell_2$ minimization based methods. 


\subsubsection{Reconstruction with different numbers of outliers}

 Now, we evaluate the recovery performance of outlier detection system under different numbers of outliers. We first fix the noise levels for MNIST and CelebA as the same as in the previous experiments. Then, for each dataset, we vary the number of outliers from 5 to 50, and measure the reconstruction error per pixel with various numbers of measurements. In these evaluations, we use $\ell_1$-minimization based algorithms {($\ell_1$ ADMM and $(\ell_1)^2$ GD)} for outlier detection and image reconstruction.

For MNIST dataset, we plot the reconstruction error performance for 5, 10, 25 and 50 outliers, respectively, in Fig. 7. One can see from Fig. 7 that as the number of outliers increases, a larger number of measurements are needed to guarantee {{successful}} image recovery. Specifically, we need at least 25 measurements to lower error rate to below 0.08 per pixel when there are 5 outliers in data, while the number of measurements should be tripled to obtain the same performance { in} the presence of 50 outliers. Next, for { the CelebA} dataset, we plot the reconstruction error performance for 5, 10, 25 and 50 outliers in Fig. 8. 

\section{Conclusions}\label{Sec:ConclusionsandFutureDirections}

In this paper, we investigated solving the outlier detection problems via a generative model approach. This new approach outperforms the $\ell_2$ minimization and traditional Lasso in both the DCT domain and Wavelet domain. The iterative alternating {{direction}} method of multipliers we proposed can {{efficiently solve the proposed nonlinear $\ell_1$ norm-based outlier detection formulation for generative model.}} Our theory shows that for both the linear neural networks and nonlinear neural networks with arbitrary {{number of}} layers, as long as they satisfy certain mild conditions, then with high probability, one can correctly detect the outliers based on generative models.

\clearpage
\begin{figure}[htb]
\begin{minipage}[b]{1.0\linewidth}
  \centering
  \centerline{\includegraphics[width=16cm]{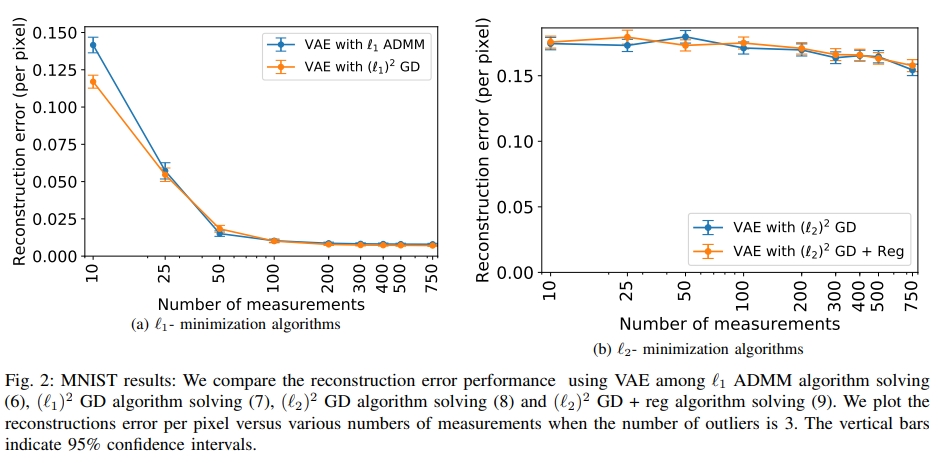}}
\end{minipage}
\end{figure} 
\begin{figure}[htb]
\begin{minipage}[b]{1.0\linewidth}
  \centering
  \centerline{\includegraphics[width=16cm]{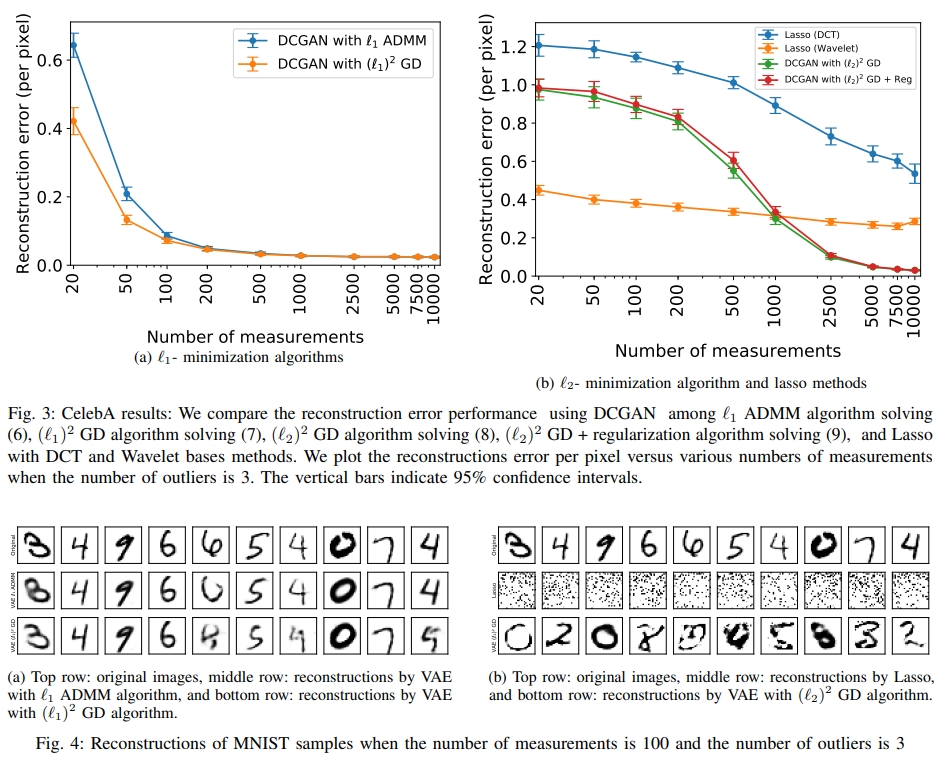}}
\end{minipage}
\end{figure} 
\begin{figure}[htb]
\begin{minipage}[b]{1.0\linewidth}
  \centering
  \centerline{\includegraphics[width=16cm]{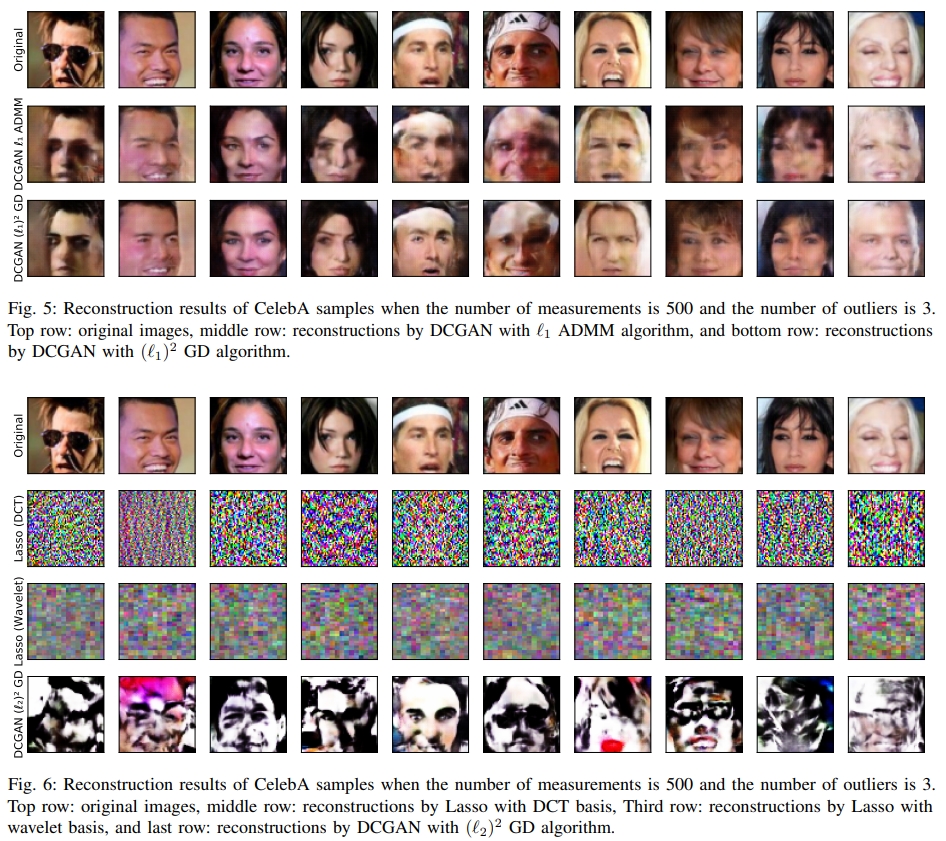}}
\end{minipage}
\end{figure} 
\begin{figure}[htb]
\begin{minipage}[b]{1.0\linewidth}
  \centering
  \centerline{\includegraphics[width=16cm]{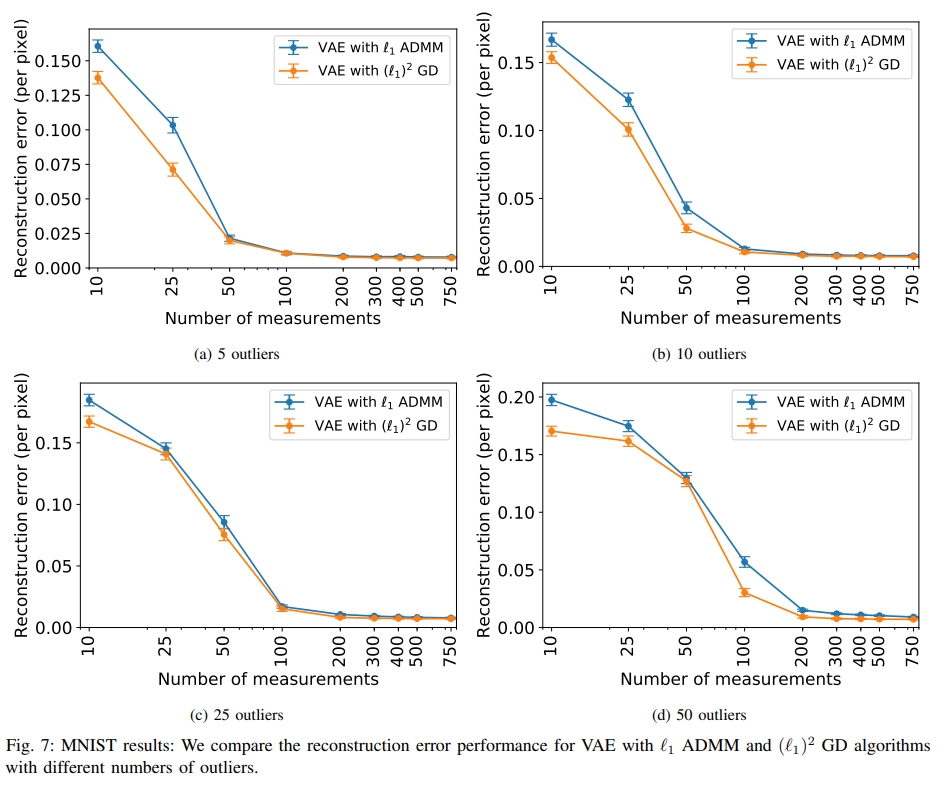}}
\end{minipage}
\end{figure} 
\begin{figure}[htb]
\begin{minipage}[b]{1.0\linewidth}
  \centering
  \centerline{\includegraphics[width=16cm]{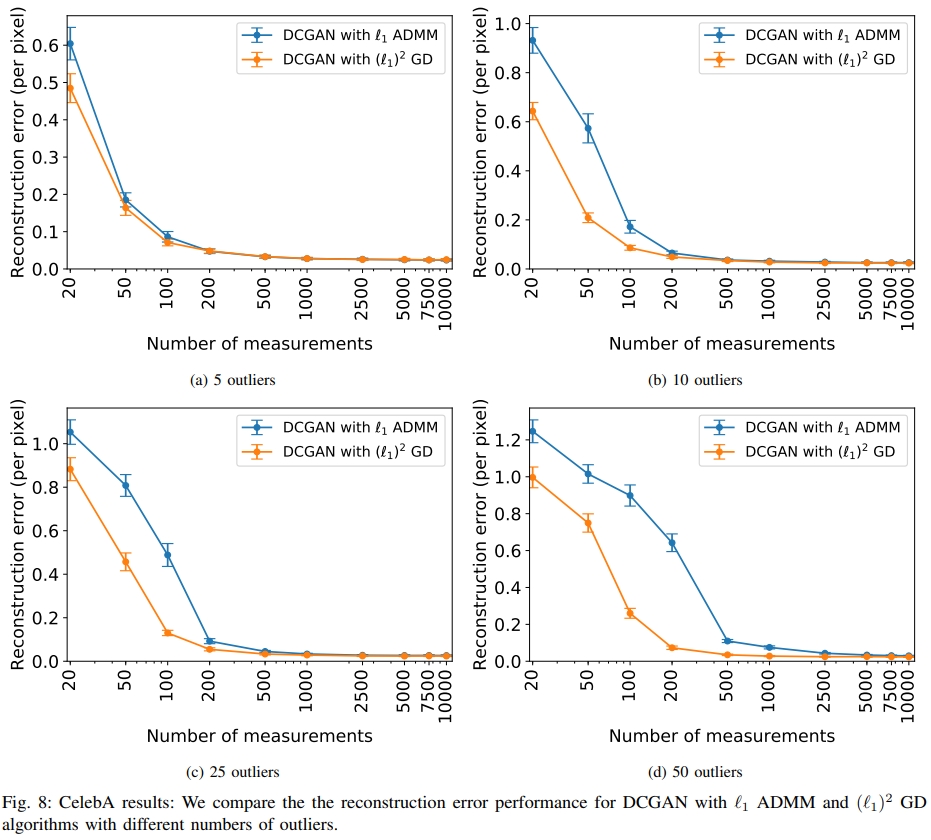}}
\end{minipage}
\end{figure} 

\bibliography{CS_ref}

\section{Appendix}
\subsection{Proof of Lemma \ref{leaky_intermediate_lowerbound}}
\begin{proof}
Define
\begin{equation}
c_{max}=\frac{1}{n}\max\limits_{\boldsymbol{v}\in \mathcal{S}}\|H\boldsymbol{v}\|_{1}.
\end{equation}
Pick a $\gamma$-net $\Sigma_{2}$ of $\mathcal{S}$ with cardinality at most $(1+2/\gamma)^{n-m}$ and $\gamma$ is a small positive number. Next, define\\
\begin{equation}
    \eta=\frac{1}{n}\max\limits_{\boldsymbol{v}\in \Sigma_{2}}\|H\boldsymbol{v}\|_{1}.
\end{equation} 
Then for every $\boldsymbol{v} \in \mathcal{S}$, there exists $\boldsymbol{v'} \in \Sigma_{2}$ such that $\|\boldsymbol{v}-\boldsymbol{v'}\|_{2}\leq \gamma$. Then,
\begin{equation}
\begin{aligned}
    \|H\boldsymbol{v}\|_{1}&\leq \|H\boldsymbol{v'}\|_{1}+\|H(\boldsymbol{v}-\boldsymbol{v'})\|_{1}\\
    &\leq \eta n+\gamma c_{max}n
\end{aligned}
\end{equation}
we therefore have
\begin{equation}
    c_{max}\leq \eta+\gamma c_{max},
\end{equation} 
which gives
\begin{equation}
    c_{max}\leq \eta/(1-\gamma).
\end{equation}
We need to first characterize $\eta$ in order to get $c_{max}$. We want to show that there exists a constant $a>E[|X|]$ such that $\|H\boldsymbol{v}\|_{1}<an$ for all $\boldsymbol{v}\in \Sigma_{2}$ with high probability, where $X \sim \mathcal{N}(0,1)$. To be specific,
\begin{equation}\label{2}
\begin{aligned}
\mathbb{P}(\eta \geq a)&=\mathbb{P}(\exists \ \boldsymbol{v} \in \Sigma_{2}\  \text{s.t.}\ \|H\boldsymbol{v}\|_{1} \geq an)\\
&\leq \sum\limits_{\boldsymbol{v}\in \Sigma_{2}}\mathbb{P}(\|H\boldsymbol{v}\|_{1}\geq an)\\
&\leq (1+2/\gamma)^{n-m}\min\limits_{s>0}e^{-sna}E[e^{s\sum\limits_{i}|H_{(i)}\boldsymbol{v}|}]\\
&=(1+2/\gamma)^{(1-\alpha)n}\min\limits_{s>0}e^{-sna}E[e^{s|X|}]^{n}\\
&=e^{((1-\alpha)\text{log}(1+2/\gamma)+\min\limits_{s>0}(\text{log}(E[e^{s|X|}])-as))n}, \ \forall s>0
\end{aligned}
\end{equation}
where the first inequality follows from the union bound and the second inequality follows from the Chernoff bound. Since the second derivative of $\text{log}(E[e^{s|X|}])-as$ in terms of $s$ is positive, so its minimum is obtained when its first derivative equals zero. In fact,
\begin{equation}
    \begin{aligned}
    0&=\frac{d[\text{log}(E[e^{s|X|}])-as]}{ds}\\
    &=\frac{d}{ds}(\text{log}(\sqrt{2/\pi}\int^{\infty}_{0}e^{sx-\frac{1}{2}x^{2}}dx)-as)\\
    &=\frac{\int^{\infty}_{0}xe^{sx-\frac{1}{2}x^{2}}dx}{\int^{\infty}_{0}e^{sx-\frac{1}{2}x^{2}}dx}-a
    \end{aligned}
\end{equation}
The solution $s$ of this is always positive since $a>E[|X|]$. Therefore, it is also the solution to $\min\limits_{s>0}(\text{log}(E[e^{s|X|}])-as)$. The exponent in (\ref{2}) tends to be negative when $a$ goes to infinity. Then, we can pick $a(\alpha,\gamma)$ large enough such that there exists some constant $c_{3}>0$, $\mathbb{P}(\eta \geq a)\leq e^{-c_{3}n}$ holds. Thus,
\begin{equation}
    \mathbb{P}(c_{max}\geq \frac{a}{1-\gamma})\leq \mathbb{P}(\frac{\eta}{1-\gamma}\geq \frac{a}{1-\gamma})\leq e^{-c_{3}n}.
\end{equation}
Next, we let
\begin{equation}
    \lambda_{max}(\alpha)=\min\limits_{\gamma}a(\alpha, \gamma)h/(1-\gamma),
\end{equation}
there exists $c_{3}$ such that with probability at least $1-e^{-c_{3}n}$, for every $\boldsymbol{d}\in \mathcal{S}$, $\|H\boldsymbol{v}\|_{1}<\lambda_{max}n/h$.
\end{proof}

\subsection{Proof of Lemma \ref{relu_lowerbound}}
\begin{proof}
Note that we have
\begin{equation}
\begin{aligned}
&\mathbb{P}(\boldsymbol{h}_{(i)}+tH'_{(i)}\boldsymbol{c'}\geq0\ |\ |H'_{(i)}\boldsymbol{c'}|=q)\\
&=\mathbb{P}(\boldsymbol{h}_{(i)}+tH'_{(i)}\boldsymbol{c'}<0\ |\ |H'_{(i)}\boldsymbol{c'}|=q)=\frac{1}{2},    
\end{aligned}
\end{equation}
and we also have 
\begin{equation}
    \mathbb{P}(\boldsymbol{h}_{(i)}+tH'_{(i)}\boldsymbol{c'}\geq0)=\mathbb{P}(\boldsymbol{h}_{(i)}+tH'_{(i)}\boldsymbol{c'}<0)=\frac{1}{2},
\end{equation}
since
$\boldsymbol{h}+tH'\boldsymbol{c'}$ is Gaussian vector.

Thus, $\boldsymbol{h}_{(i)}+tH'_{(i)}\boldsymbol{c'}\geq0$ and $|H'_{(i)}\boldsymbol{c'}|=q$ are independent. Then we have the right derivative of $\|\sigma(\boldsymbol{h}+tH'\boldsymbol{c}')-\sigma(\boldsymbol{h})\|_{1}$ equal to 
\begin{equation}
    \sum_{i=1}^{n}\mathcal{I}(\boldsymbol{h}_{i}+tH'_{(i)}\boldsymbol{c'}\geq0)\cdot|H'_{(i)}\boldsymbol{c'}|=\sum_{i=1}^{n}\mathcal{B}_{i}|H'_{(i)}\boldsymbol{c'}|,
\end{equation}
where $\mathcal{I}$ is the indicator function, $H'_{(i)}$ is the $i^{th}$ row of the matrix $H'$, and $\mathcal{B}_{i}$ $\sim$ Bernoulli$(\frac{1}{2})$ is Bernoulli variable.

In fact, by Chernoff bound, we have
\begin{equation}\label{4}
\begin{aligned}
&\mathbb{P}(\sum_{i=1}^{n}|\mathcal{B}_{i}H'_{(i)}\boldsymbol{c'}| \leq bn)\\
\leq &e^{bns}E[e^{-s\sum\limits_{i}|\mathcal{B}_{i}H'_{(i)}\boldsymbol{c'}|}]\\
=&e^{(\text{log}(E[e^{-s|Y|}])+bs)n}, \ \forall s>0
\end{aligned}
\end{equation}
Here the random variable $Y$ is the product of a Bernoulli random variable and a Gaussian random variable. We can write $Y=ZX$ where $X\sim \mathcal{N}(0,1)$ and $Z\sim$ Bernoulli$(\frac{1}{2})$. Then, if $y<0$, we have:
\begin{equation}
\begin{aligned}
&\mathbb{P}(Y\leq y)\\
=&\mathbb{P}(XZ\leq y|z=1)\cdot \mathbb{P}(z=1)+\mathbb{P}(XZ\leq y|z=0)\cdot \mathbb{P}(z=0)\\
=&\frac{1}{2}\mathbb{P}(X\leq y)+\frac{1}{2}\mathbb{P}(0\leq y)\\
=&\frac{1}{2}\mathbb{P}(X\leq y)\\
=&\frac{1}{2}\int_{-\infty}^{y}\frac{1}{\sqrt{2\pi}}e^{-\frac{x^{2}}{2}}dx.
\end{aligned}
\end{equation}
Similarly, when $y\geq0$, we have:
\begin{equation}
\begin{aligned}
&\mathbb{P}(Y\leq y)\\
=&\mathbb{P}(XZ\leq y|z=1)\cdot \mathbb{P}(z=1)+\mathbb{P}(XZ\leq y|z=0)\cdot \mathbb{P}(z=0)\\
=&\frac{1}{2}\mathbb{P}(X\leq y)+\frac{1}{2}\mathbb{P}(0\leq y)\\
=&\frac{1}{2}\mathbb{P}(X\leq y)+\frac{1}{2}\\
=&\frac{1}{2}+\frac{1}{2}\int_{-\infty}^{y}\frac{1}{\sqrt{2\pi}}e^{-\frac{x^{2}}{2}}dx.
\end{aligned}
\end{equation}
Therefore, the probability density function of $Y$ when $Y\neq 0$ is $f(y)=\frac{1}{2}\cdot\frac{1}{\sqrt{2\pi}}e^{-\frac{y^{2}}{2}}$ and $\mathbb{P}(Y=0)=\frac{1}{2}$.
Note that
\begin{equation}
    \begin{aligned}
    E[e^{-s|Y|}]&=\frac{1}{2}e^{-s\cdot 0}+\frac{1}{2}\cdot2\int_{0}^{\infty}e^{-sy}\cdot\frac{1}{2}\cdot\frac{1}{\sqrt{2\pi}}e^{-\frac{y^{2}}{2}}dy\\
    &=\frac{1}{2}+\frac{1}{2\sqrt{2\pi}}\int^{\infty}_{0}e^{-sy}e^{-\frac{y^2}{2}}dy\\
    &=\frac{1}{2}+\frac{1}{2\sqrt{2\pi}s}\int^{\infty}_{0}e^{-x}e^{-\frac{1}{2}(x/s)^{2}}dx\\
    &\leq\frac{1}{2}+\frac{1}{2\sqrt{2\pi}s}\int^{\infty}_{0}e^{-x}dx\\
    &=\frac{1}{2}+\frac{1}{2\sqrt{2\pi}s}
    \end{aligned}
\end{equation}
where we do the substitution $x=sy$ in the third equation.
Also, from the third equation, when $s>1$, we will have
\begin{equation}
    E[e^{-s|Y|}]\geq \frac{1}{2}+\frac{1}{2\sqrt{2\pi}s}\int^{\infty}_{0}e^{-x-\frac{1}{2}x^{2}}dx
\end{equation}
By combining the above, we get 
\begin{equation}
    E[e^{-s|Y|}]=\frac{1}{2}+C(s)\cdot \frac{1}{s}.
\end{equation}
where $C(s)$ is a positive number as a function of $s$, and
\begin{equation}
0.13\approx\frac{\int^{\infty}_{0}e^{-x-\frac{1}{2}x^{2}}dx}{2\sqrt{2\pi}}\leq C(s) \leq \frac{1}{2\sqrt{2\pi}}
\end{equation}
We further let
\begin{equation}
    b=\frac{1}{2s}.
\end{equation}
Denote
\begin{equation}
\begin{aligned}
    \kappa&=-\text{log}(E[e^{-s|Y|}])-bs\\
    &=-\text{log}(\frac{1}{2}+C(s)\cdot \frac{1}{s})-\frac{1}{2}. 
\end{aligned}
\end{equation}

Then
\begin{equation}
    \mathbb{P}(\sum_{i=1}^{n}|\mathcal{B}_{i}H'_{(i)}\boldsymbol{c'}| \leq bn)\leq e^{-\kappa n}.
\end{equation}
A possible set of values for parameters is $s=2, b=\frac{1}{4}$

For this choice of $s$ and $b$, we verified that $\kappa>0$. Therefore, there exists $\kappa>0$ such that $\mathbb{P}(\sum_{i=1}^{n}|\mathcal{B}_{i}H'_{(i)}\boldsymbol{c'}| \leq bn=\frac{n}{4})\leq e^{-\kappa n}$. We take $c_{7}=\kappa$ to end this proof.
\end{proof}
\end{document}